\documentclass[11pt]{article}

\usepackage[preprint]{acl}

\usepackage{times}
\usepackage{latexsym}

\usepackage[T1]{fontenc}
\usepackage[utf8]{inputenc}

\usepackage{microtype}

\usepackage{inconsolata}

\usepackage{graphicx}
\usepackage{amsmath}
\usepackage{booktabs}
\usepackage{colortbl}
\usepackage[table]{xcolor}
\definecolor{headercolor}{RGB}{219,217,226}
\usepackage{caption}
\usepackage{pifont}
\usepackage{xcolor}
\usepackage{amsfonts}
\usepackage[most]{tcolorbox}
\tcbuselibrary{breakable}

\newcommand{\cmark}{\textcolor{green!70!black}{\ding{51}}}
\newcommand{\xmark}{\textcolor{red!70!black}{\ding{55}}}

\title{ToolPRMBench: Evaluating and Advancing Process Reward Models for Tool-using Agents}

\author{
  Dawei Li\textsuperscript{\ding{171}}, 
  Yuguang Yao\textsuperscript{\ding{169}},  
  Zhen Tan\textsuperscript{\ding{171}}, 
  Huan Liu\textsuperscript{\ding{171}}, 
  Ruocheng Guo\textsuperscript{\ding{169}} \\
  \textsuperscript{\ding{171}}Arizona State University, 
  \textsuperscript{\ding{169}}Intuit AI Research \\
  {\tt \{daweili5, ztan36, huanliu\}@asu.edu}\\
  {\tt \{yuguang\_yao,ruocheng\_guo\}@intuit.com}
}

\begin{document}
\maketitle
\begin{abstract}
% Large language models (LLMs) are increasingly used as agents that interact with external tools to solve complex, multi-step tasks. However, monitoring their tool use remains challenging. Early mistakes can propagate through long sequences of such interaction. Yet, the prevailing outcome-based monitor provides limited insight into intermediate errors. 
Reward-guided search methods have demonstrated strong potential in enhancing tool-using agents by effectively guiding sampling and exploration over complex action spaces. As a core design, those search methods utilize process reward models (PRMs) to provide step-level rewards, enabling more fine-grained monitoring. However, there is a lack of systematic and reliable evaluation benchmarks for PRMs in tool-using settings. In this paper, we introduce ToolPRMBench, a large-scale benchmark specifically designed to evaluate PRMs for tool-using agents. ToolPRMBench is built on top of several representative tool-using benchmarks and converts agent trajectories into step-level test cases. Each case contains the interaction history, a correct action, a plausible but incorrect alternative, and relevant tool metadata. We respectively utilize offline sampling to isolate local single-step errors and online sampling to capture realistic multi-step failures from full agent rollouts. A multi-LLM verification pipeline is proposed to reduce label noise and ensure data quality. We conduct extensive experiments across large language models, general PRMs, and tool-specialized PRMs on ToolPRMBench. The results reveal clear differences in PRM effectiveness and highlight the potential of specialized PRMs for tool-using. Code and data will be released at \url{https://github.com/David-Li0406/ToolPRMBench}.
\end{abstract}

\section{Introduction}
In recent years, Large Language Models (LLMs) have achieved remarkable success, exhibiting exceptional performance across a wide range of applications—from complex reasoning and multilingualism to highly specialized fields~\citep{openai2024gpt4technicalreport,li2025system,ren2025deepseek,yang2025quantifying,tan2024large,wang2024bpo,zhang2025guilomoallocatingexpertnumber,jiang2024memorization,zhang2025find,qin2025survey,zhang-etal-2025-shifcon,jiang2025drp,qwen2025qwen25technicalreport}.
Complementing this progress, a rapidly growing area of focus is the deployment of LLMs as tool-using agents, which interact with external tools such as APIs, databases, and execution environments to solve complex, multi-step tasks~\cite{shen2024llm,huang2024metatool,tan2024interpreting,lee2025knowledge,zhao2025scale,yu2025chain,xu2025learning,chang2025treereview,tan2025prospect}.
% Large Language Models (LLMs) are increasingly deployed as tool-using agents, where they interact with external tools such as APIs, databases, and execution environments to solve complex, multi-step tasks~\cite{shen2024llm,huang2024metatool,tan2024interpreting,lee2025knowledge,zhao2025scale,yu2025chain,chang2025treereview,tan2025prospect}. 
This paradigm significantly extends the capability of LLMs beyond pure text generation. However, effective tool-using remains challenging. As highlighted in recent work, even strong models frequently fail due to early mistakes that propagate through long sequences, while final outcome-based evaluation provides limited insight into where the reasoning process goes wrong \cite{yao2022react,liang2025sws,chae2025web,shahroz-etal-2025-agents}.

\begin{table}[t]
\centering
\small
\renewcommand{\arraystretch}{0.9}
\setlength{\tabcolsep}{4pt}
\begin{tabular}{lccc}
\toprule
\textbf{Benchmark} & \textbf{Step-level} & \textbf{Agent} & \textbf{Tool-using} \\
\midrule
Agent-RewardBench \\ \cite{men-etal-2025-agent} & \xmark & \cmark & \textcolor{gray}{Web-only} \\
AgentRewardBench \\ \cite{lu2025agentrewardbench} & \xmark & \cmark & \textcolor{gray}{Web-only} \\
WebRewardBench  \\ \cite{chae2025web}  & \cmark & \cmark & \textcolor{gray}{Web-only} \\
PRMBench \\ \cite{song-etal-2025-prmbench}         & \cmark & \xmark & \xmark \\
\midrule
\textbf{ToolPRMBench} & \textbf{\cmark} & \textbf{\cmark} & \textbf{\textcolor{green!70!black}{Diverse APIs}} \\
\bottomrule
\end{tabular}
\caption{ToolPRMBench is the first benchmark that supports step-level evaluation for interactive agents with diverse tool APIs.}
\label{tab:benchmark_comparison}
\end{table}

Motivated by these challenges, recent research has increasingly explored reward-guided search as a way to improve tool-using agents. Instead of committing to a single trajectory, reward-guided methods perform searching or sampling over multiple candidate actions or plans. In tool-using and web agent settings, methods such as best-of-$n$ or Monte Carlo tree search have shown promising results~\cite{chae2025web,agarwal2025toolrm,zhang2025symbiotic}. A central component of these approaches is a Process Reward Model (PRM)~\cite{snell2024scaling,zhao2025genprm}, which provides step-level feedback to guide exploration and prune incorrect trajectories early. Compared to outcome-only rewards, PRMs offer finer-grained signals that are better aligned with the long-horizon tool-using tasks. However, despite their growing importance, evaluating PRMs for tool-using remains a challenging task. Tool-using, as an agent task, is characterized by long interactions and a large, structured action space, where errors can emerge at many intermediate steps and propagate over time; as a result, existing PRM benchmarks designed for general reasoning~\cite{song-etal-2025-prmbench} or web agents~\cite{men-etal-2025-agent,lu2025agentrewardbench} may not be directly applicable or sufficiently effective in tool-using scenarios. Moreover, existing PRM designs vary widely, ranging from LLM-as-a-judge methods~\cite{li2025generation} to general PRMs~\cite{yang2024qwen2} or agent-~\cite{chae2025web} and tool-specialized PRMs. However, there is no unified benchmark to systematically evaluate their effectiveness in tool-using settings.

To address this gap, we introduce \textbf{ToolPRMBench}, a large-scale benchmark designed specifically for evaluating process reward models for tool-using agents (Figure~\ref{tab:benchmark_comparison}). ToolPRMBench is constructed on top of several representative tool-using benchmarks, covering diverse environments such as information-seeking, multi-step reasoning, and interactive tool execution. 
% In total, the benchmark contains a large number of step-level preference pairs, where each sample corresponds to a single decision step in an agent trajectory. 
Each sample in ToolPRMBench consists of the interaction history, a correct action, an incorrect but plausible alternative, and associated tool metadata. To build the dataset, we combine offline sampling, which isolates local single-step errors around golden trajectories, and online sampling, which captures realistic multi-step failures from full agent rollouts. These candidate samples are further verified through a multi-LLM filtering pipeline to ensure label reliability. This construction process results in a diverse and challenging benchmark that enables fine-grained evaluation of PRMs at the decision-step level.

Through extensive experiments on ToolPRMBench across a total of 17 large language models, general PRMs, and tool-specialized PRMs, we observe clear and consistent performance differences across model types. Our results further show that scaling model size and general capabilities benefit tool process reward modeling, while reinforcement learning demonstrates strong potential for improving robustness and generalization in tool-using PRMs. In addition, we conduct a series of analyses on ToolPRMBench, including meta-evaluation, data synthesis, cost analysis, and case studies, providing further insights to guide future research on reward modeling for tool-using agents.

In summary, the contribution in this work is threefold:
\begin{itemize}
    \item First, we propose ToolPRMBench, a large-scale benchmark carefully constructed for systematic evaluation of PRMs in tool-using settings.
    \item Second, we benchmark a series of LLMs, general PRMs, and tool-using specialized PRMs in ToolPRMBench, building a comprehensive PRMs leaderboard in the tool-using scenario.
    \item Finally, we conduct further analysis, including meta-evaluation and cost analysis, providing findings and insights for future reward-guided trajectory searching in tool-using.
\end{itemize}

\section{Related Work}

\subsection{Tool-using Benchmarks}
Tool-using agents extend LLMs by enabling them to interact with external tools or APIs rather than only producing text. This capability is necessary because many real-world tasks cannot be solved by language generation alone and require actual execution of external actions to achieve correct outcomes.
Several benchmarks have been proposed to evaluate tool-using capabilities~\cite{farn2023tooltalk,grattafiori2024llama,wang2024gta,lu2025toolsandbox}. Some focus on whether a model can utilize tools where appropriate~\cite{huang2024metatool,ning2024wtu}, while others measure multi-hop or structured tool invocation, such as ToolHop~\cite{ye2025toolhop} and MCP-RADAR~\cite{gao2025mcp}. Additionally, evaluation frameworks like T-Eval and Trajectory-Bench decompose tool utilization into sub-processes (instruction following, planning, reasoning, retrieval, etc.) to provide fine-grained analysis beyond end-to-end task success~\cite{chen2024t,he2025traject}.
Despite progress, many existing benchmarks focus solely on final task success, rather than fine-grained and step-level accuracy. Therefore, we propose ToolPRMBench, a benchmark designed to evaluate PRM across diverse tool-using trajectories and scenarios.

\begin{figure*}
    \centering
    \includegraphics[width=1\linewidth]{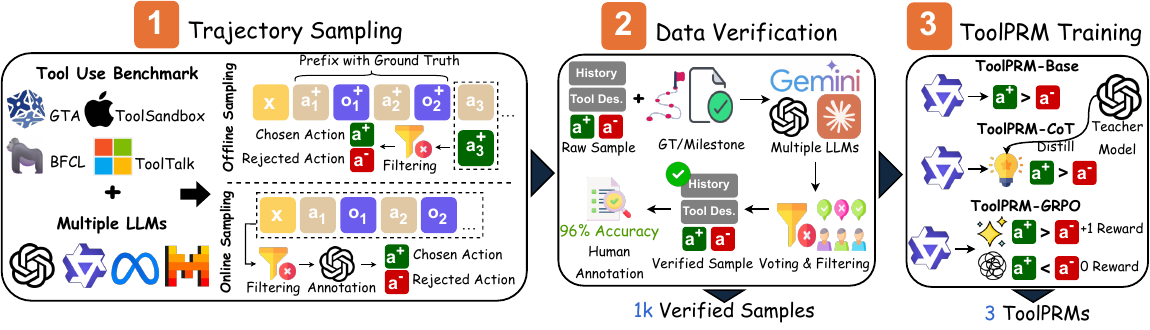}
    \caption{The overview pipeline of ToolPRMBench, including trajectory sampling, data verification \& filtering, and ToolPRM training.}
    \label{fig:placeholder}
\end{figure*}

\subsection{Reward-guided Search and Reward Modeling}
Reward-guided search refers to strategies that improve model performance at test time by sampling and searching over multiple candidate actions.
In general tasks, reward-guided search relies on learned or heuristic reward models to evaluate candidate outputs, including outcome-based reward models trained from human or AI preferences~\cite{stiennon2020learning}, classifier-style evaluators or value functions for reranking~\cite{cobbe2021training}, and self-consistency or majority-vote signals derived from multiple sampled trajectories~\cite{wangself}.
For long-horizon settings such as agents and tool-using systems, however, sparse outcome rewards are often insufficient, and successful reward-guided search critically depends on process reward modeling and accurate step-level rewards that can guide intermediate decisions and credit assignment~\cite{uesato2022solving,lightman2023let}.
Such step-wise guidance is especially important when agents must balance reasoning depth, action selection, and external tool costs over extended trajectories~\cite{zhou2024language,chae2025web, jiang2026scribe}.
Motivated by recent progress in this direction, we introduce ToolPRMBench to evaluate the effectiveness of PRMs in tool-using.
% Reward-guided search refers to strategies that improve model performance at test time by sampling and searching over multiple candidate actions. In the context of general LLMs, this includes techniques such as sampling multiple candidates~\cite{}, reranking with external evaluators~\cite{}, and adaptive compute allocation to improve output quality under a fixed model size~\cite{}. Recently, inference-time scaling has also been studied in the context of agents that interact with external environments and tools. As agents incur both token generation costs and action costs (e.g., invocation of external tools), effective scaling must balance reasoning depth and resource usage. For example, TUMIX proposes a multi-agent mixture technique that uses parallel reasoning paths and iterative refinement to improve tool-using outcomes within inference cost constraints~\cite{}. These methods illustrate that inference-time strategies for agentic systems must consider both language generation and action selection. Among these approaches, PRMs are particularly promising, as they provide step-level guidance for efficient inference-time scaling in agentic settings~\cite{}. Motivated by recent progress in this direction, we introduce ToolPRMBench to evaluate the effectiveness of PRMs in tool-using.

\section{ToolPRMBench}

ToolPRMBench is a benchmark for PRMs in tool-using agents. It aggregates trajectories using multiple existing tool-using benchmarks and converts them into step-wise test samples, enabling systematic evaluation of whether a given PRM can distinguish correct actions from incorrect ones at each decision step. Following the design of the previous PRMBench~\cite{chae2025web} for web agents, ToolPRMBench focuses on process-level correctness, rather than just final task success. ToolPRMBench is constructed on top of four representative tool-using benchmarks, including ToolTalk~\cite{farn2023tooltalk}, GTA~\cite{wang2024gta}, BFCL~\cite{patilberkeley}, and ToolSandbox~\cite{lu2025toolsandbox}. These benchmarks cover diverse environments for tool-using agents such as information-seeking, multi-step reasoning, and interactive tool execution. 

% For each benchmark instance, we first obtain a high-quality golden trajectory that correctly completes the user instruction. We then generate alternative actions using large language models (LLMs) and convert them into preference pairs after verification. The overall construction pipeline consists of collecting golden trajectories, generating rejected actions through trajectory sampling, filtering noisy samples, and finally building step-level preference data.

Each sample in ToolPRMBench corresponds to a single decision step in a sequence of interactions between the agent and the environment. Let $x$ denote the user instruction, and $\tau = ((a_1, o_1), (a_2, o_2), \ldots, (a_T, o_T))$ represent a trajectory of actions of an agent, where $a_t$ denotes the action at step $t$ (i.e., a structured tool call) and $o_t$ is the corresponding observation returned by the environment. The interaction history at step $t$ is defined as $h_t = (x, a_1, o_1, \ldots, a_{t-1}, o_{t-1})$. Then, a ToolPRMBench sample is a tuple $(h_t, a_t^{+}, a_t^{-}, m_t)$, where $a_t^{+}$ is the chosen (correct) action, $a_t^{-}$ is the rejected (incorrect) action, and $m_t$ contains metadata such as the available tool description.

\subsection{Trajectory Sampling}

To collect rejected actions, we adopt two trajectory sampling strategies that complement each other: offline and online sampling. Both strategies leverage the golden trajectory or milestone, but differ in the level of flexibility the model has when generating actions.

Offline sampling constrains the model to follow the golden trajectory prefix and only samples an alternative action at a specific step. This setting focuses on local errors and isolates single-step mistakes. In contrast, online sampling allows the model to generate an entire trajectory from the beginning freely, which naturally leads to multi-step, correlated errors. Online sampling better reflects realistic agent failures but is harder to analyze. In ToolPRMBench, we choose the sampling strategy based on the characteristics of the original benchmark and its evaluation protocol.

\noindent\textbf{Offline Sampling.}
Let $\tau^{\star} = ((a_1^{\star}, o_1^{\star}), \ldots, (a_T^{\star}, o_T^{\star}))$ be the golden trajectory for instruction $x$. At step $t$, the golden history is $h_t^{\star} = (x, a_1^{\star}, o_1^{\star}, \ldots, a_{t-1}^{\star}, o_{t-1}^{\star})$. We query a tool-using policy $\pi$ to sample an action $\tilde{a}_t \sim \pi(\cdot \mid h_t^{\star})$. Importantly, the environment is not updated with $\tilde{a}_t$, and subsequent steps always follow the golden trajectory.

To construct samples in offline sampling, we compare $\tilde{a}_t$ with the golden action $a_t^{\star}$. If the two actions are semantically equivalent, the step is discarded. Otherwise, we create a candidate sample $(h_t^{\star}, a_t^{+} = a_t^{\star}, a_t^{-} = \tilde{a}_t)$. The comparison is performed using task-specific rules that examine the tool name and key arguments. Offline sampling, therefore, produces localized deviations around a correct history and provides clean supervision for step-level error detection.

\noindent\textbf{Online Sampling.}
Offline sampling cannot capture error propagation across multiple steps. To address this limitation, we additionally collect data using online sampling in interactive benchmarks such as BFCL and ToolSandbox. Given instruction $x$, the policy $\pi$ generates a full trajectory $\hat{\tau} = ((\hat{a}_1, \hat{o}_1), \ldots, (\hat{a}_{\hat{T}}, \hat{o}_{\hat{T}}))$ by interacting with the environment. Each generated trajectory is evaluated using the benchmark’s outcome-based metric, resulting in a binary success signal $s(\hat{\tau}) \in \{0,1\}$. We retain only failed trajectories with $s(\hat{\tau}) = 0$.

To identify the erroneous step in a failed trajectory, we employ an LLM-based annotation process. The annotator LLM is given the user instruction, the generated trajectory, the metadata, and a golden reference (either the full golden trajectory or task milestones). It is asked to identify the first incorrect step index $t_{\text{err}}$ and to propose a corrected action $\bar{a}_{t_{\text{err}}}$. The history is then defined as $\hat{h}_{t_{\text{err}}} = (x, \hat{a}_1, \hat{o}_1, \ldots, \hat{a}_{t_{\text{err}}-1}, \hat{o}_{t_{\text{err}}-1})$, and the resulting preference pair is $(\hat{h}_{t_{\text{err}}}, a_t^{+} = \bar{a}_{t_{\text{err}}}, a_t^{-} = \hat{a}_{t_{\text{err}}})$. This procedure converts trajectory-level failures into step-level supervision, making it suitable for PRM evaluation.

\subsection{Data Verification}

Both offline and online sampling can introduce noise. In offline sampling, the golden trajectory is not necessarily the only valid solution; therefore, some sampled actions may be incorrectly labeled as rejected. In online sampling, LLM-based annotation may misidentify the error step or propose an incorrect correction. To mitigate these issues, we apply a multi-LLM verification and filtering pipeline.

For each candidate sample $(h_t, a_t^{+}, a_t^{-}, m_t)$, we query three powerful LLMs (GPT-5, Gemini-3-flash and Claude-4.5-haiku) to independently judge whether $a_t^{+}$ is strictly better than $a_t^{-}$ given the history. Each model provides a binary judgment, and we aggregate the results using majority voting. Samples that receive unanimous positive votes are retained, while samples unanimously rejected are discarded. For borderline cases with mixed votes, we perform additional human verification on a subset of samples. This multi-judge strategy significantly reduces label noise and improves the reliability of the test set in ToolPRMBench. To further validate the effectiveness of multi-LLM verification, we randomly sample 100 samples from all LLM-verified results and observe a 96\% agreement with human judgments.

\subsection{ToolPRM Training}
\label{sec:toolprm-training}

We describe the training objectives of different ToolPRM variants. Each training instance is represented as
\begin{equation}
(h_t, \tilde{a}_t^{(1)}, \tilde{a}_t^{(2)}, m_t),
\end{equation}
where $\{\tilde{a}_t^{(1)}, \tilde{a}_t^{(2)}\}$ is a random permutation of the chosen action $a_t^{+}$ and the rejected action $a_t^{-}$ to avoid position bias~\cite{li2025generation}. We consider the following three training methods.

\noindent\paragraph{ToolPRM-Base.}
ToolPRM-Base is trained to directly predict which candidate action should be selected. Given the input $(h_t, \tilde{a}_t^{(1)}, \tilde{a}_t^{(2)}, m_t)$, the model outputs an action label $y_t \in \{1, 2\}$, where $y_t$ indicates the position of the chosen action in the permuted candidate list. Specifically, $y_t = 1$ if the chosen action $a_t^{+}$ appears at position $\tilde{a}_t^{(1)}$, and $y_t = 2$ otherwise. The model is trained using supervised fine-tuning (SFT) with a cross-entropy loss:
\begin{equation}
\mathcal{L}_{\text{SFT}} = - \log p(y_t \mid h_t, \tilde{a}_t^{(1)}, \tilde{a}_t^{(2)}, m_t).
\end{equation}

\noindent\paragraph{ToolPRM-CoT.}
ToolPRM-CoT extends the base model by explicitly modeling the reasoning process. For each input $(h_t, \tilde{a}_t^{(1)}, \tilde{a}_t^{(2)}, m_t)$, the model is trained to generate a reasoning sequence $r_t$ followed by the action label $y_t$. The reasoning sequence $r_t$ used during training is from a larger teacher model. Both $r_t$ and the final action label $y_t$ are optimized jointly using supervised fine-tuning:
\begin{equation}
\mathcal{L}_{\text{SFT}} = - \log p(r_t, y_t \mid h_t, \tilde{a}_t^{(1)}, \tilde{a}_t^{(2)}, m_t).
\end{equation}

\noindent\paragraph{ToolPRM-GRPO.}
ToolPRM-GRPO further improves the model using reinforcement learning with Group Relative Policy Optimization (GRPO)~\cite{shao2024deepseekmath}. For each case, the policy samples multiple reasoning--action pairs $(r_t, y_t)$ conditioned on $(h_t, \tilde{a}_t^{(1)}, \tilde{a}_t^{(2)}, m_t)$. We define a binary reward function:
\begin{equation}
R(y_t) =
\begin{cases}
1, & \text{if } y_t \text{ corresponds to the position of } a_t^{+}, \\
0, & \text{otherwise}.
\end{cases}
\end{equation}

The training objective is to maximize the expected reward:
\begin{equation}
\mathbb{E}_{(r_t,y_t) \sim p_\theta(\cdot \mid h_t,\tilde{a}_t^{(1)},\tilde{a}_t^{(2)},m_t)}[R(y_t)].
\end{equation}
This reinforcement learning stage encourages the model to refine both the reasoning and action selection behavior beyond supervised fine-tuning.

% \subsection{Error Injection}

% Constructing large-scale preference data with real executions is expensive. To reduce the cost of training PRMs while maintaining distributional consistency with the test set, we adopt a metadata-guided error injection strategy to synthesize training data.

% We first analyze the verified ToolPRMBench test set to identify common error types, such as incorrect tool selection, wrong arguments, ignoring prior observations, or premature termination. We also estimate the empirical distribution of error types and their positions within trajectories. This information is stored as metadata and used to guide data synthesis.

% For each golden trajectory, we sample an error type according to the metadata distribution. We then prompt an LLM to inject a natural and realistic error at the selected step while keeping the rest of the trajectory unchanged. The injected action $\tilde{a}_t$ forms a rejected action, paired with the original golden action $a_t^{\star}$ as the chosen action, resulting in a synthetic preference pair $(h_t^{\star}, a_t^{+} = a_t^{\star}, a_t^{-} = \tilde{a}_t)$.

% Crucially, the LLM used for error injection only receives high-level metadata and golden trajectories, and never observes actual test samples. This design prevents data leakage while allowing us to generate scalable, realistic training data that closely matches the evaluation distribution of ToolPRMBench.

\subsection{Dataset Statistics}

\begin{figure}
    \centering
    \includegraphics[width=1\linewidth]{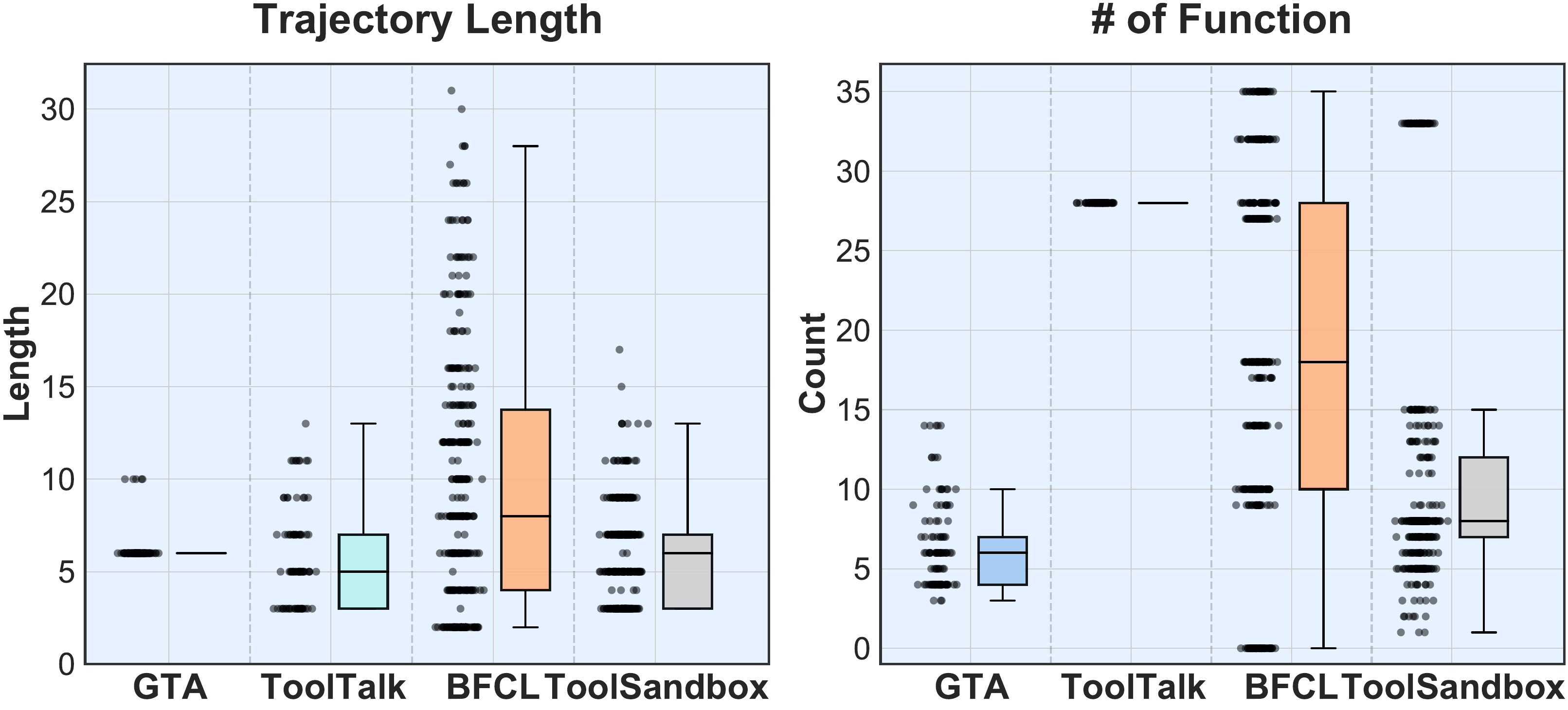}
    \caption{Statistics of trajectory length and number of functions in ToolPRMBench.}
    \label{fig:trajectory_function}
\end{figure}

\begin{figure}
    \centering
    \includegraphics[width=1\linewidth]{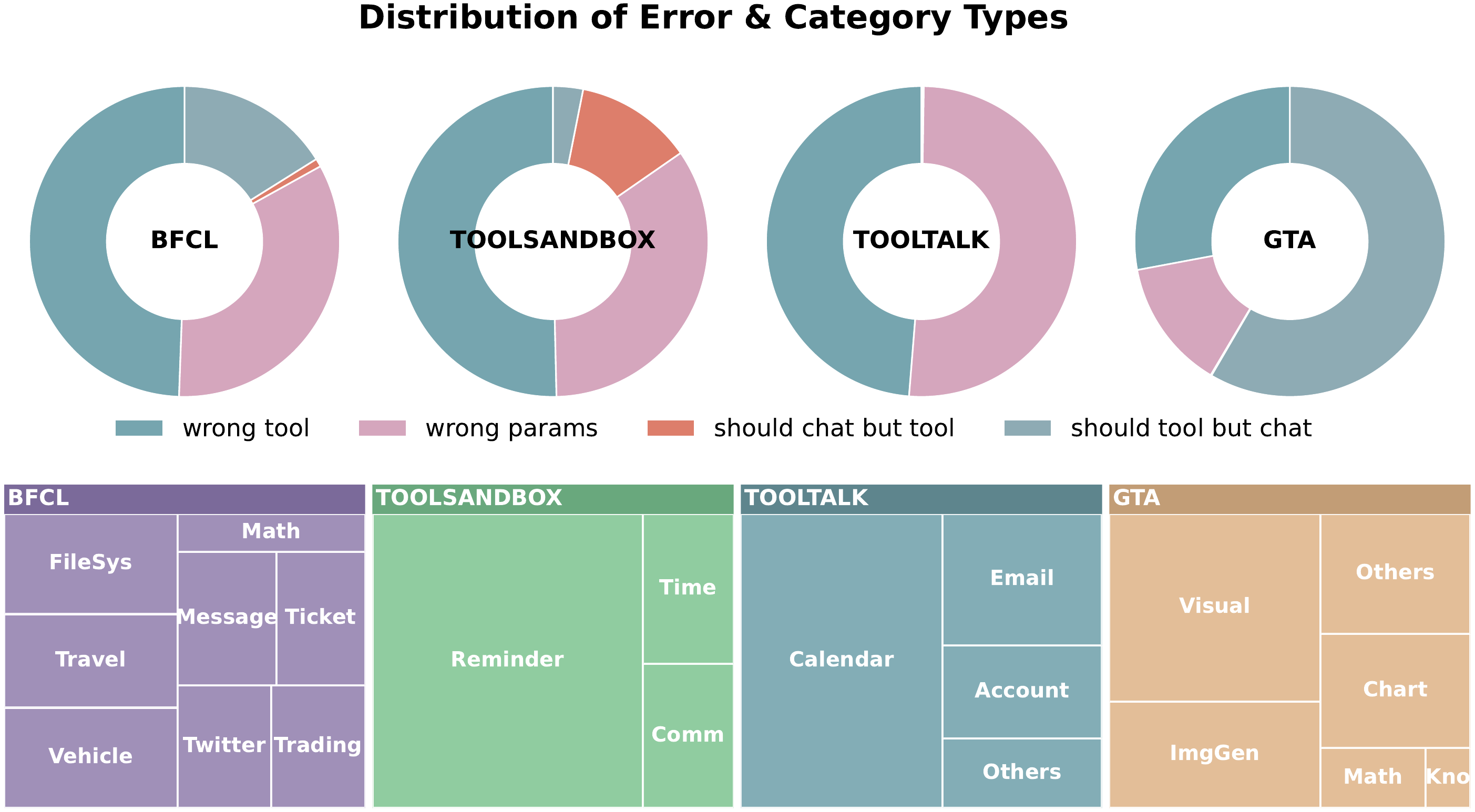}
    \caption{Statistics of error and category distribution in ToolPRMBench.}
    \label{fig:error_category}
\end{figure}

There are 984 samples in the ToolPRMBench in total. Figure~\ref{fig:trajectory_function} and \ref{fig:error_category} present the distribution of error, category, trajectory length, and function number in four subsets of ToolPRMBench. We found that most trajectories have moderate lengths, covering both short interactions and more complex multi-step processes. This distribution allows the test set to probe model performance across varying levels of task difficulty without being dominated by trivial cases. Additionally, the data includes common failure modes, such as incorrect action selection, incorrect arguments, and improper use of tools, as well as natural language responses. These errors appear at different positions along trajectories, supporting a comprehensive assessment of a model’s ability to detect and rank incorrect decisions throughout the interaction. Overall, the statistics indicate that ToolPRMBench is diverse and challenging, making it suitable for fine-grained evaluation of step-level decision quality. More statistics and details of the ToolPRMBench collection can be found in Appendix~\ref{app:ToolPRMBench}.

\section{Experiment}

\subsection{Experiment Setting}

\noindent\paragraph{Models.} We benchmark a series of LLMs in our ToolPRMBench, including (1). API-Based LLMs: GPT-5~\cite{openai2025gpt5}, Claude-4.5-haiku~\cite{anthropic2025sonnet45} and Gemini-2.5-flash~\cite{comanici2025gemini}. (2). Open-source LLMs, including models in Qwen3~\cite{yang2025qwen3} and LLaMA-3~\cite{grattafiori2024llama} families. (3). General PRMs for math and web navigation, including WebShepHerd-8B~\cite{chae2025web}, Qwen2.5-Math-7B~\cite{yang2024qwen2}, Llemma-7b-prm~\cite{sun2024easy} and Math-shepherd~\cite{wang2024math}. (4). Tool-using specialized PRMs, including ToolPRM-Base, ToolPRM-CoT, and ToolPRM-GRPO.

\noindent\paragraph{Implementation Details.} ToolPRM is trained on parts of the BFCL and ToolSandbox subset in ToolPRMBench, with the training and testing ratio to be 7:3. To prevent data contamination, we ensure that all ToolPRMBench samples derived from the same instruction are assigned exclusively to either the training set or the testing set. All ToolPRM variants are trained on top of Qwen-3-4B. For ToolPRM-CoT, the reasoning supervision is distilled from GPT-5-mini. Model training is conducted using LLaMA-Factory~\cite{zheng2024llamafactory} and TRL~\cite{vonwerra2022trl}, and inference is performed with the vLLM backend engine. More details about experiment implementation can be found in Appendix~\ref{app:Implementation}.

\subsection{Main Result}

\begin{table*}[h!]
\small
\renewcommand{\arraystretch}{0.9}
\centering
\rowcolors{1}{gray!8}{white}
\begin{tabular}{lccccc}
\toprule
\rowcolor{headercolor}
\textbf{Model} & \textbf{GTA} & \textbf{ToolTalk} & \textbf{BFCL} & \textbf{ToolSandbox} & \textbf{AVG} \\
\midrule

\multicolumn{6}{c}{\textit{API-Based LLMs}} \\
GPT-5~\cite{openai2025gpt5}            & 87.3 & \underline{82.5} & 44.1 & \underline{83.7} & 74.4 \\
Claude-4.5-haiku~\cite{anthropic2025sonnet45}       & \underline{91.5} & \textbf{93.0} & 45.9 & \textbf{70.0} & \textbf{75.1} \\
Gemini-2.5-flash~\cite{comanici2025gemini}       & 90.1 & 86.7 & 40.8 & 75.3 & 73.2 \\
\midrule

\multicolumn{6}{c}{\textit{Open-Source LLMs}} \\
Qwen3-1.7B~\cite{yang2025qwen3}       & 50.8 & 50.0 & 36.7 & 38.1 & 43.9 \\
Qwen3-4B~\cite{yang2025qwen3}         & 63.5 & 66.2 & 30.1 & 37.8 & 49.4 \\
Qwen3-8B~\cite{yang2025qwen3}         & 66.1 & 68.6 & 35.2 & 45.0 & 53.7 \\
Qwen3-14B~\cite{yang2025qwen3}        & 74.6 & 80.1 & 35.2 & 62.1 & 63.0 \\
LLaMA-3-3B-Instruct~\cite{grattafiori2024llama}       & 40.7 & 41.9 & 40.5 & 50.7 & 43.4 \\
LLaMA-3-8B-Instruct~\cite{grattafiori2024llama}       & 42.4 & 50.0 & 47.2 & 39.7 & 44.8 \\
LLaMA-3-70B-Instruct~\cite{grattafiori2024llama}      & 65.3 & 70.1 & 43.2 & 36.0 & 53.6 \\
\midrule

\multicolumn{6}{c}{\textit{General PRMs}} \\
WebShepHerd-8B~\cite{chae2025web}   & 52.0 & 64.0 & 37.5 & 43.4 & 49.2 \\
Qwen2.5-Math-7B~\cite{yang2024qwen2}  & 29.7 & 69.4 & 36.9 & 67.0 & 50.8 \\
Llemma-7b-prm~\cite{sun2024easy}    & 41.5 & 64.1 & 45.0 & 60.8 & 52.8 \\
Math-shepherd~\cite{wang2024math}    & 59.3 & 33.6 & 53.0 & 57.7 & 50.9 \\
\midrule

\multicolumn{6}{c}{\textit{Tool-using PRMs}} \\
ToolPRM-Base     & 38.1 & \underline{65.1} & 47.7 & \underline{77.7} & 57.1 \\
ToolPRM-CoT      & \underline{55.1} & 56.9 & \underline{57.7} & \textbf{83.0} & \underline{63.2} \\
ToolPRM-GRPO     & \textbf{84.7} & \textbf{73.3} & \textbf{86.4} & 70.0 & \textbf{78.6} \\

\bottomrule
\end{tabular}
\caption{Main experiment result in ToolPRMBench. Best result in each subset is \textbf{bold}; second best is \underline{underlined}.}
\label{tab:converted_styled}
\end{table*}

\begin{figure*}[t!]
    \centering
    \includegraphics[width=1\linewidth]{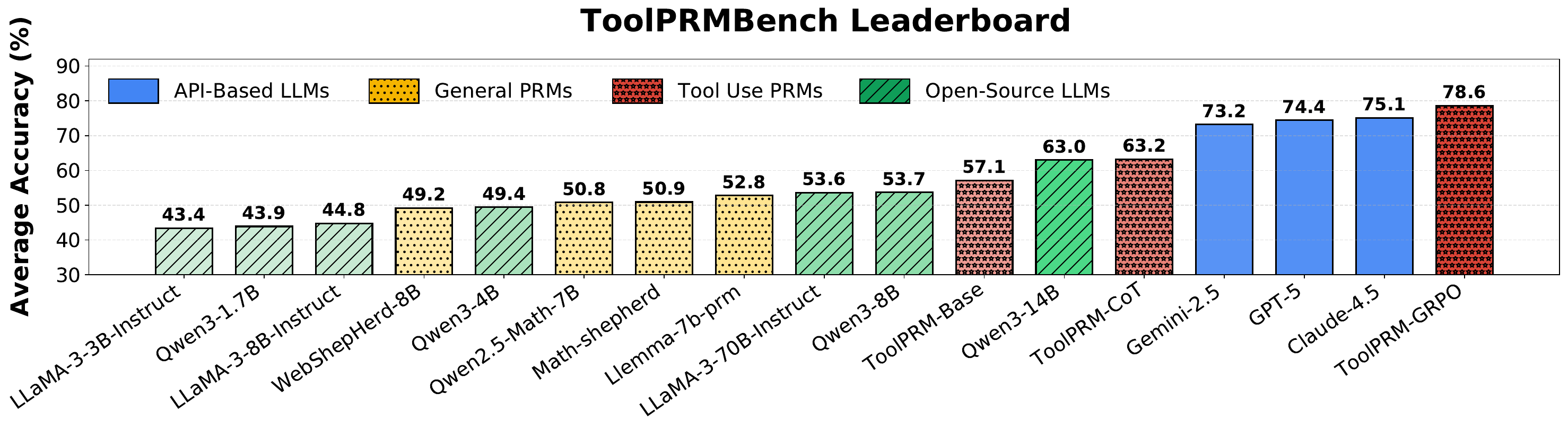}
    \caption{ToolPRM leaderboard on 17 LLMs.}
    \label{fig:leaderboard}
\end{figure*}

\noindent\paragraph{Overall comparison across model categories.}
Table~\ref{tab:converted_styled} and the leaderboard in Figure~\ref{fig:leaderboard} provide a comprehensive comparison of various model types on ToolPRMBench. A clear performance hierarchy emerges across model families. API-based LLMs consistently achieve the strongest overall results, ranking at the top across almost all subsets. This suggests that large-scale training and strong general reasoning abilities continue to be highly effective for process-level evaluation in tool-using scenarios.

Tool-specialized PRMs also demonstrate strong performance. In particular, ToolPRM-GRPO achieves the best average accuracy among all non-API models, outperforming even the API-based LLMs. ToolPRM-CoT and ToolPRM-Base further show that reward models explicitly trained for tool-using substantially outperform both open-source LLMs and general-purpose PRMs. In contrast, open-source LLMs and general PRMs exhibit noticeably weaker performance. Many of these models struggle to exceed 55\% average accuracy, suggesting that PRMs trained for math reasoning or web navigation do not directly transfer to tool-using process evaluation. Overall, these results highlight the importance of tool-specific supervision when designing effective process reward models.

\begin{figure}[h!]
    \centering
    \includegraphics[width=1\linewidth]{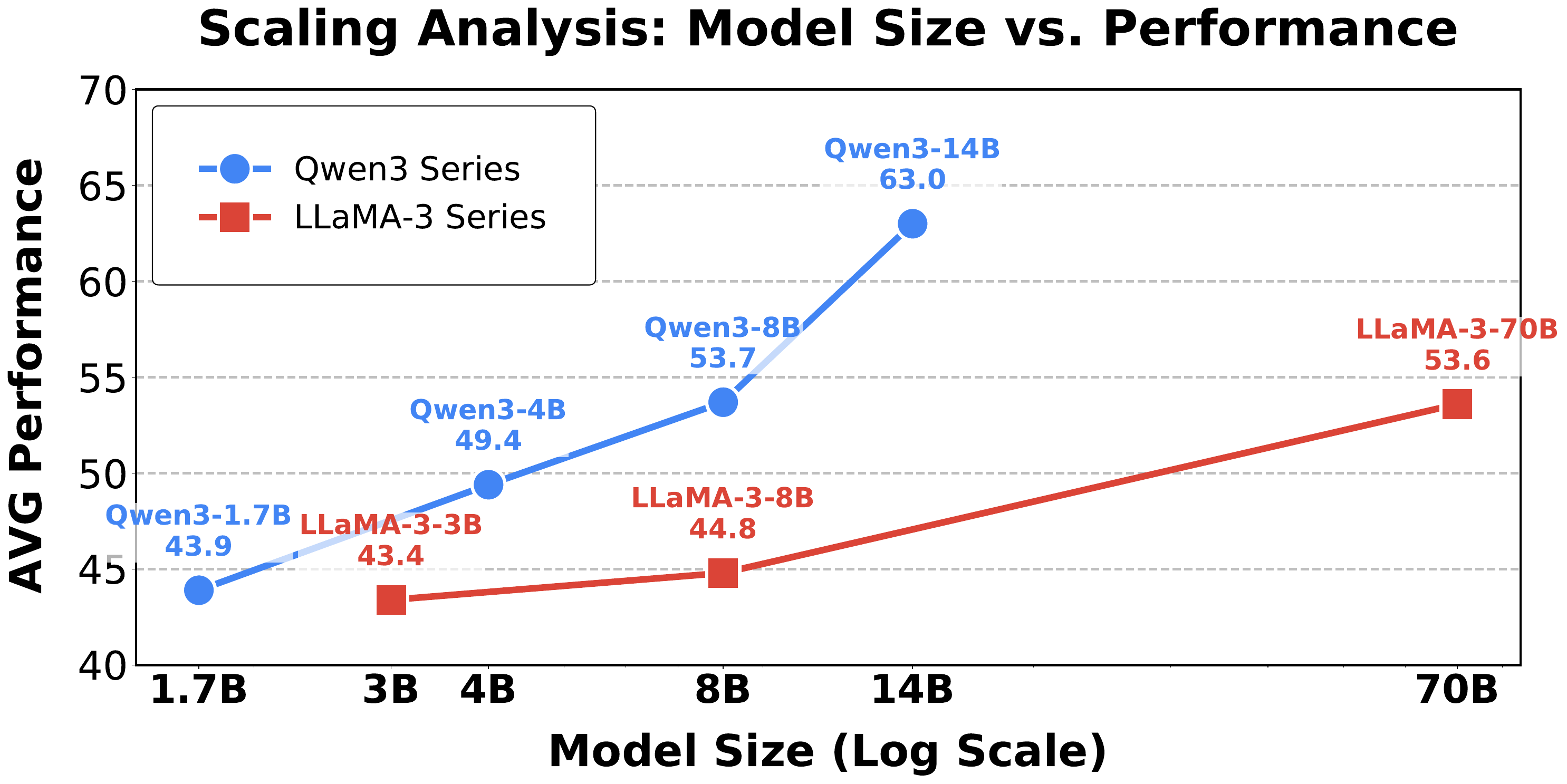}
    \caption{Scaling analysis result on Qwen3 and LLaMA-3.}
    \label{fig:leaderboard}
\end{figure}

\paragraph{Scaling behavior of base models.}
Figure~\ref{fig:leaderboard} reports the scaling analysis over the Qwen3 and LLaMA-3 model families. A clear positive trend can be observed between model size and performance on ToolPRMBench. This result suggests that improvements in general model capacity, such as reasoning ability and instruction following, are beneficial for tool process reward modeling. However, scaling alone is not sufficient. Even the largest open-source models still lag behind ToolPRMs by a significant margin. This gap suggests that while model size and general performance are beneficial, specialized training remains crucial for achieving strong performance in tool-using PRMs.

\begin{figure}[h!]
    \centering
    \includegraphics[width=1\linewidth]{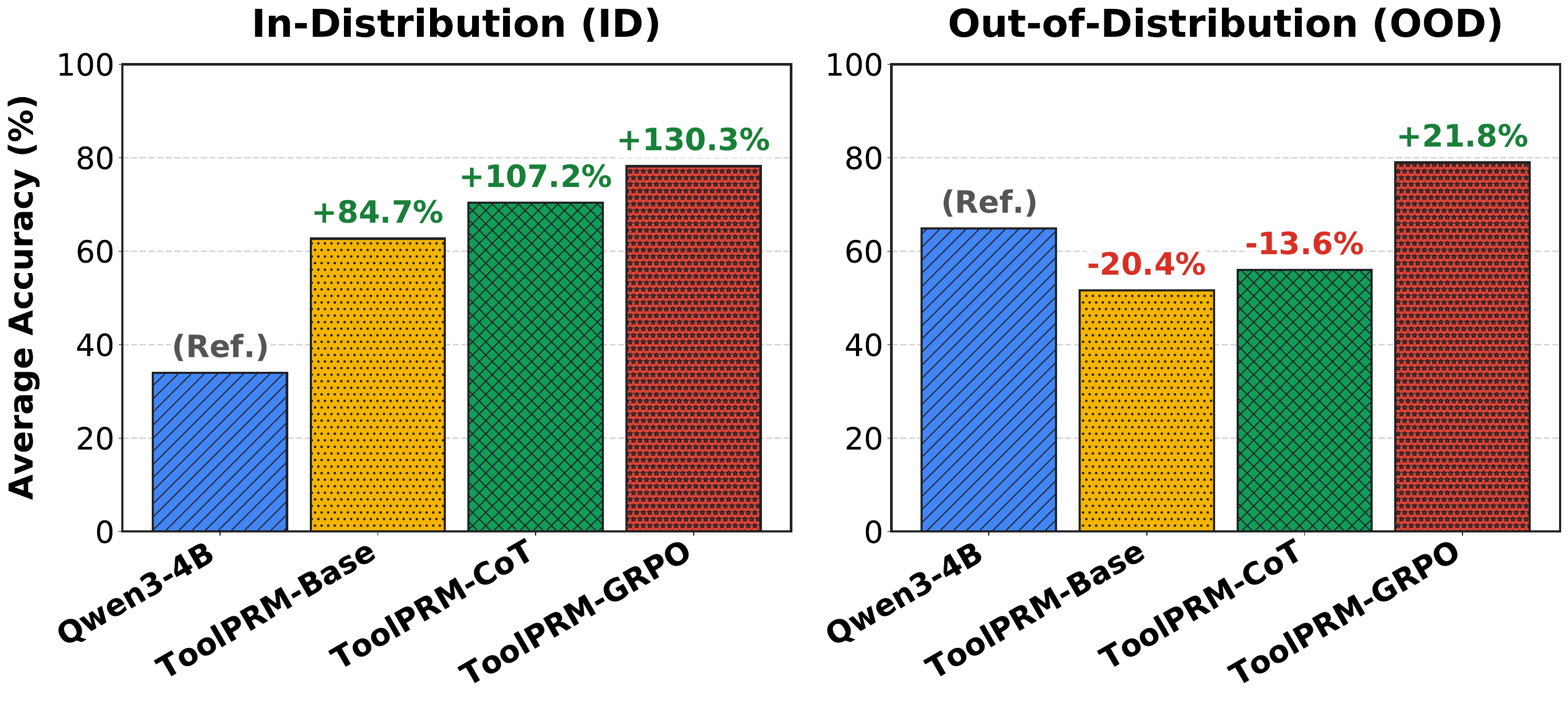}
    \caption{ToolPRMs' results in ID and OOD settings.}
    \label{fig:id_vs_ood}
\end{figure}

\paragraph{In-distribution vs.\ out-of-distribution generalization.}
Figure~\ref{fig:id_vs_ood} compares three ToolPRM variants under in-distribution (ID) and out-of-distribution (OOD) evaluation settings. ToolPRM-Base and ToolPRM-CoT, both trained using supervised fine-tuning, show clear improvements in the ID setting. However, their performance drops substantially in the OOD setting, with relative decreases of 20.4\% and 13.6\%, respectively. This behavior suggests that SFT-based methods are prone to overfitting and may rely on distribution-specific patterns that do not generalize well.

In contrast, ToolPRM-GRPO achieves consistent gains in both ID and OOD evaluations, with a 21.8\% improvement in the OOD setting. This result indicates that reinforcement learning encourages more robust decision boundaries and reduces reliance on spurious correlations. Overall, these findings suggest that RL-based training is a promising direction for ToolPRM learning, particularly when generalization is required beyond the training distribution.

\section{Further Analysis}

\subsection{Meta-Evaluation}

\begin{figure}[h!]
    \centering
    \includegraphics[width=1\linewidth]{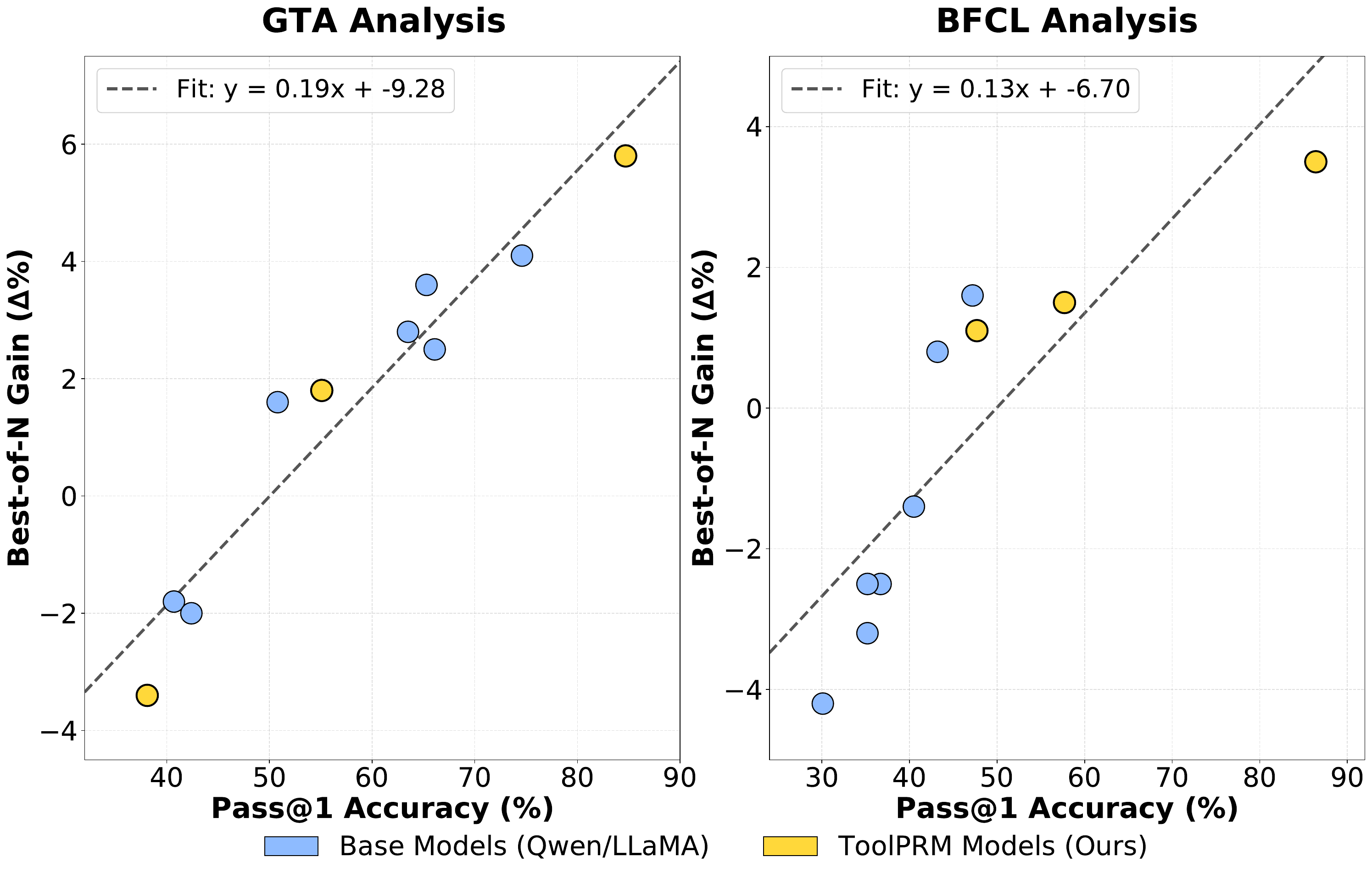}
    \caption{Meta-evaluation of ToolPRMBench on GTA and BFCL.}
    \label{fig:correlation}
\end{figure}

To examine whether ToolPRMBench reflects PRM performance under realistic and dynamic conditions, we conduct a meta-evaluation on GTA and BFCL. Specifically, we use different models as reward functions to guide best-of-$n$ search with $n=8$, and measure the resulting performance gains. Figure~\ref{fig:correlation} illustrates the correlation between ToolPRMBench accuracy and the effectiveness of reward-guided search.

We observe a strong positive correlation on both benchmarks. Models that perform well on ToolPRMBench consistently yield larger gains during reward-guided search, indicating that ToolPRMBench is a reliable proxy for PRM effectiveness in inference-time decision making. At the same time, models with poor ToolPRMBench performance (accuracy below 50\%) often yield negative gains. In these cases, using such models as reward functions actively harms performance, as they tend to misguide exploration and amplify incorrect trajectories.

\subsection{Can Synthetic Data Improve ToolPRMs?}

Collecting high-quality pairwise data for ToolPRM training is costly and challenging. To alleviate this issue, we explore a simple data synthesis strategy that constructs preference pairs by directly inserting incorrect actions into ground-truth trajectories following~\cite{wang2024self}. This approach avoids additional rollouts and reduces annotation cost while preserving the overall task structure.

\begin{figure}
    \centering
    \includegraphics[width=1\linewidth]{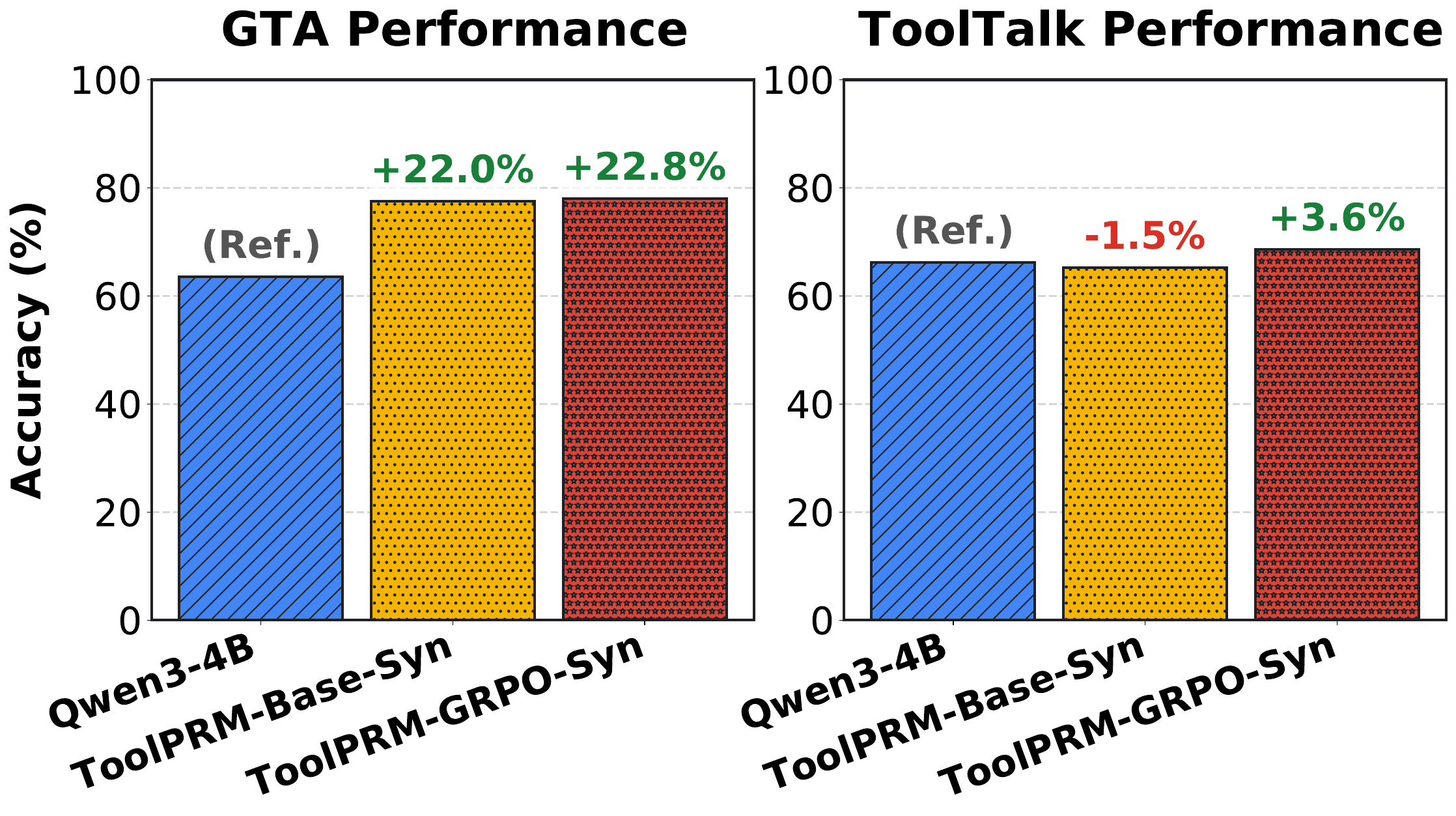}
    \caption{ToolPRMs' results with synthetic data on GTA and ToolTalk.}
    \label{fig:synthesis_analysis}
\end{figure}

We apply this data synthesis strategy to GTA and ToolTalk, and train both ToolPRM-Base and ToolPRM-GRPO using the synthesized data. As shown in Figure~\ref{fig:synthesis_analysis}, synthetic data leads to substantial improvements on GTA, with both models achieving over 22\% relative gains. However, the effect on ToolTalk is much weaker. ToolPRM-Base-Syn slightly degrades performance, whereas ToolPRM-GRPO-Syn shows only marginal improvement.

These results suggest that synthetic data is a promising direction. However, its effectiveness strongly depends on the task and environment. Designing more realistic and diverse synthetic errors remains a significant challenge, and more advanced synthesis strategies are needed for broader applicability.

\subsection{Cost Analysis}

\begin{figure}
    \centering
    \includegraphics[width=1\linewidth]{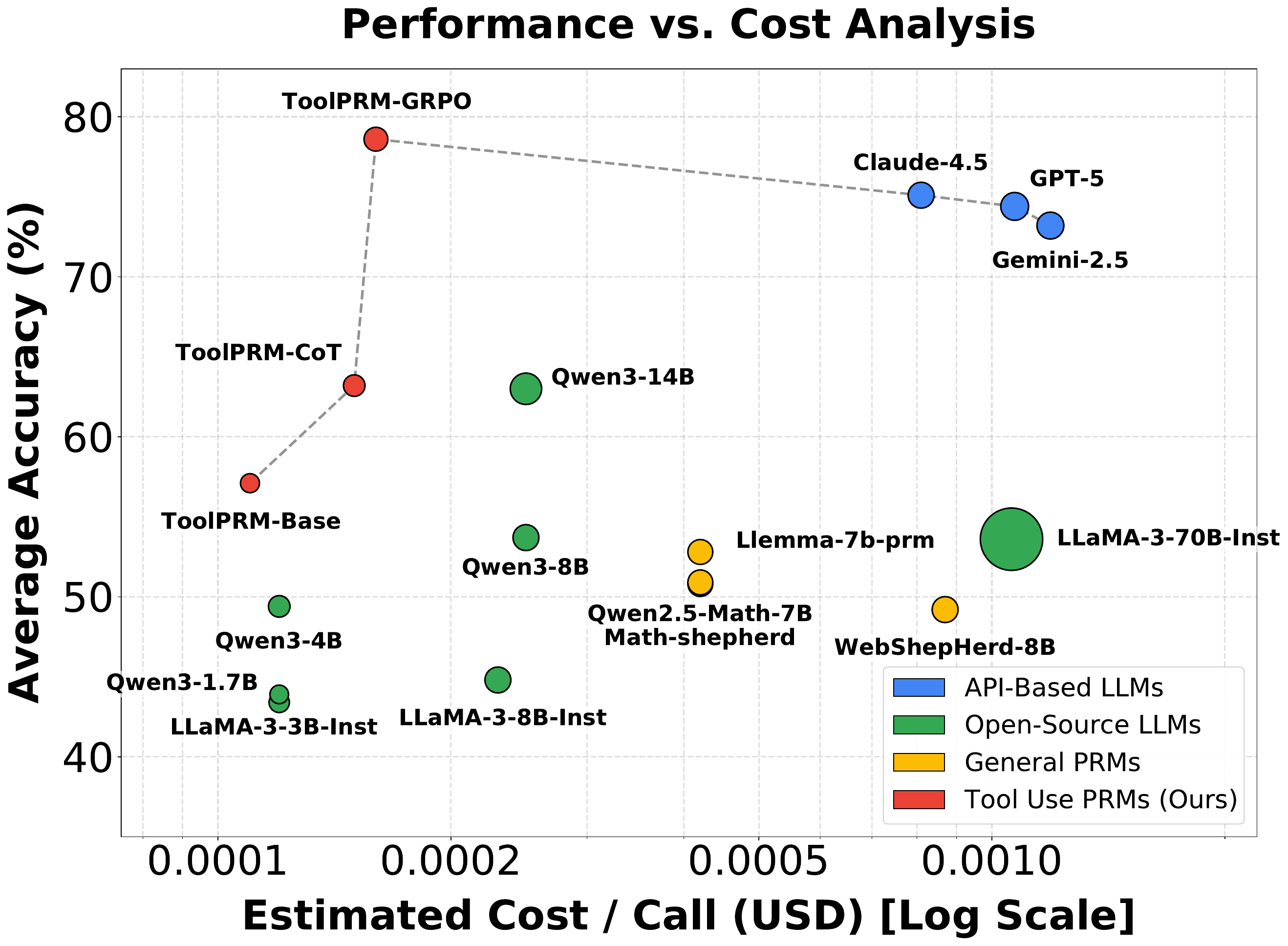}
    \caption{Cost analysis result across 4 types of models.}
    \label{fig:perf_vs_cost}
\end{figure}
We further analyze the trade-off between performance and inference cost across different model categories, as shown in Figure~\ref{fig:perf_vs_cost}. For API-based LLMs, we estimate the per-call cost using the official API pricing provided by the respective service. For open-source LLMs, we adopt the pricing provided by Together.ai\footnote{https://www.together.ai/} under a unified inference setup. For ToolPRMs and other general PRMs, we use the cost of their corresponding base model.

The results reveal a clear performance--cost trade-off. API-based LLMs achieve strong performance on ToolPRMBench, but their inference costs are significantly higher than those of other models. In contrast, ToolPRMs operate at a much lower cost while still delivering competitive accuracy, substantially outperforming open-source LLMs and general PRMs under similar or even lower budgets. This finding further supports the practicality and promise of tool-specific PRMs for reward-guided inference-time scaling in real-world agent systems.

\subsection{Case Study}

\begin{figure}
    \centering
    \includegraphics[width=1\linewidth]{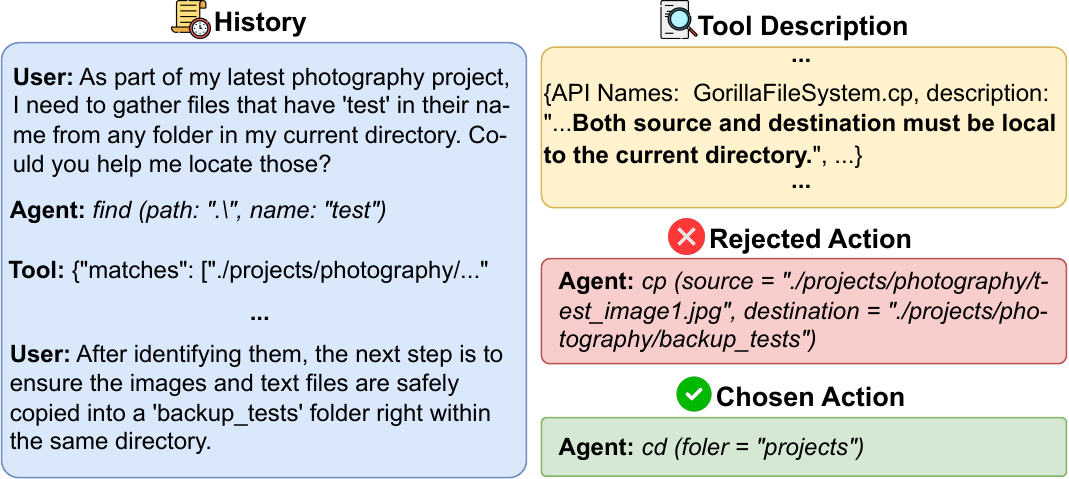}
    \caption{A case study from the BFCL subset.}
    \label{fig:case_study}
\end{figure}

Figure~\ref{fig:case_study} presents a representative example from the BFCL subset in ToolPRMBench, illustrating a common process-level error in tool-using agents. The user first requests to locate files containing ``test'' in their name, which the agent correctly accomplishes using the \texttt{find} tool. After the relevant files and directories are identified, the user asks to copy the images and text files into the \texttt{backup\_tests} directory. Although the rejected action correctly reflects the high-level intent of copying files, it violates the tool specification of \texttt{cp}, which requires both the source and destination to be local to the current working directory. By directly providing file paths, the agent ignores the implicit state constraint imposed by the file system tools. In contrast, the chosen action first changes the working directory using \texttt{cd}, which is a necessary precondition for performing valid copy operations under the given tool interface. This example highlights that errors in tool-using agents often arise not from misunderstanding user intent, but from failing to satisfy low-level tool constraints and state transitions.

\section{Conclusion}
We present ToolPRMBench, a large-scale benchmark for evaluating process reward models in tool-using agent settings. It's collected across diverse Tool-using benchmarks, combining both offline and online trajectory sampling. Through extensive evaluation on ToolPRMBench, we observe clear and consistent differences in tool process reward modeling across model families. The results indicate that increased model scale and stronger general capabilities are beneficial, but they are not sufficient without specialized training. Reinforcement learning plays a key role in improving robustness and generalization. We further conduct a set of complementary analyses on ToolPRMBench, including meta-evaluation with reward-guided search, studies on distribution shift, synthetic data, and efficiency, offering practical insights for future research on reliable and scalable reward modeling in tool-using agents.

\section*{Limitations}
Despite the promising results demonstrated by ToolPRMBench, our study has several limitations that provide directions for future research. First, although recent studies have highlighted the potential of inference-time scaling methods with RL~\cite{qian2025toolrl}, we did not conduct extensive evaluations of these search-based strategies on our benchmark. Given the constraints on time and available computing resources, our experiments primarily focus on the intrinsic discriminative ability of PRMs rather than their end-to-end impact under heavy training budgets. Future work could explore more efficient RL algorithms to bridge this gap. Second, the current construction of ToolPRMBench is based on a selected set of representative tool-using benchmarks. Recently, new datasets and protocols based on the Model Context Protocol (MCP) have emerged, offering more standardized ways for agents to interact with diverse tools. However, incorporating these MCP-based environments often involves high data collection costs and complex environment setups. Due to budget considerations during the data collection phase, we did not include these specific datasets in the current version of our benchmark. Expanding the scope to include MCP-compatible tools would likely enhance the diversity and real-world applicability of the evaluation.

\bibliography{custom}

\begin{thebibliography}{63}
\providecommand{\natexlab}[1]{#1}

\bibitem[{Agarwal et~al.(2025)Agarwal, Abdelaziz, Basu, Unuvar, Lastras, Rizk, and Kapanipathi}]{agarwal2025toolrm}
Mayank Agarwal, Ibrahim Abdelaziz, Kinjal Basu, Merve Unuvar, Luis~A Lastras, Yara Rizk, and Pavan Kapanipathi. 2025.
\newblock Toolrm: Outcome reward models for tool-calling large language models.
\newblock \emph{arXiv preprint arXiv:2509.11963}.

\bibitem[{{Anthropic}(2025)}]{anthropic2025sonnet45}
{Anthropic}. 2025.
\newblock \href {https://www.anthropic.com/news/claude-sonnet-4-5} {Introducing claude sonnet 4.5}.
\newblock \url{https://www.anthropic.com/news/claude-sonnet-4-5}.
\newblock Official announcement of Claude Sonnet 4.5, a next-generation AI model optimized for coding, agents, and sustained reasoning tasks.

\bibitem[{Chae et~al.(2025)Chae, Kim, Cho, Kim, Moon, Hwangbo, Lim, Kim, Hwang, Gwak et~al.}]{chae2025web}
Hyungjoo Chae, Sunghwan Kim, Junhee Cho, Seungone Kim, Seungjun Moon, Gyeom Hwangbo, Dongha Lim, Minjin Kim, Yeonjun Hwang, Minju Gwak, and 1 others. 2025.
\newblock Web-shepherd: Advancing prms for reinforcing web agents.
\newblock \emph{arXiv preprint arXiv:2505.15277}.

\bibitem[{Chang et~al.(2025)Chang, Li, Zhang, Kong, Wu, So, Guo, Zhu, and Wong}]{chang2025treereview}
Yuan Chang, Ziyue Li, Hengyuan Zhang, Yuanbo Kong, Yanru Wu, Hayden Kwok-Hay So, Zhijiang Guo, Liya Zhu, and Ngai Wong. 2025.
\newblock Treereview: A dynamic tree of questions framework for deep and efficient llm-based scientific peer review.
\newblock In \emph{Proceedings of the 2025 Conference on Empirical Methods in Natural Language Processing}, pages 15662--15693.

\bibitem[{Chen et~al.(2024)Chen, Du, Zhang, Liu, Liu, Zheng, Zhuo, Zhang, Lin, Chen et~al.}]{chen2024t}
Zehui Chen, Weihua Du, Wenwei Zhang, Kuikun Liu, Jiangning Liu, Miao Zheng, Jingming Zhuo, Songyang Zhang, Dahua Lin, Kai Chen, and 1 others. 2024.
\newblock T-eval: Evaluating the tool utilization capability of large language models step by step.
\newblock In \emph{Proceedings of the 62nd Annual Meeting of the Association for Computational Linguistics (Volume 1: Long Papers)}, pages 9510--9529.

\bibitem[{Cobbe et~al.(2021)Cobbe, Kosaraju, Bavarian, Chen, Jun, Kaiser, Plappert, Tworek, Hilton, Nakano et~al.}]{cobbe2021training}
Karl Cobbe, Vineet Kosaraju, Mohammad Bavarian, Mark Chen, Heewoo Jun, Lukasz Kaiser, Matthias Plappert, Jerry Tworek, Jacob Hilton, Reiichiro Nakano, and 1 others. 2021.
\newblock Training verifiers to solve math word problems.
\newblock \emph{arXiv preprint arXiv:2110.14168}.

\bibitem[{Comanici et~al.(2025)Comanici, Bieber, Schaekermann, Pasupat, Sachdeva, Dhillon, Blistein, Ram, Zhang, Rosen et~al.}]{comanici2025gemini}
Gheorghe Comanici, Eric Bieber, Mike Schaekermann, Ice Pasupat, Noveen Sachdeva, Inderjit Dhillon, Marcel Blistein, Ori Ram, Dan Zhang, Evan Rosen, and 1 others. 2025.
\newblock Gemini 2.5: Pushing the frontier with advanced reasoning, multimodality, long context, and next generation agentic capabilities.
\newblock \emph{arXiv preprint arXiv:2507.06261}.

\bibitem[{Farn and Shin(2023)}]{farn2023tooltalk}
Nicholas Farn and Richard Shin. 2023.
\newblock Tooltalk: Evaluating tool-usage in a conversational setting.
\newblock \emph{arXiv preprint arXiv:2311.10775}.

\bibitem[{Gao et~al.(2025)Gao, Xie, Zhai, Ma, and Shen}]{gao2025mcp}
Xuanqi Gao, Siyi Xie, Juan Zhai, Shiqing Ma, and Chao Shen. 2025.
\newblock Mcp-radar: A multi-dimensional benchmark for evaluating tool use capabilities in large language models.
\newblock \emph{arXiv preprint arXiv:2505.16700}.

\bibitem[{Grattafiori et~al.(2024)Grattafiori, Dubey, Jauhri, Pandey, Kadian, Al-Dahle, Letman, Mathur, Schelten, Vaughan et~al.}]{grattafiori2024llama}
Aaron Grattafiori, Abhimanyu Dubey, Abhinav Jauhri, Abhinav Pandey, Abhishek Kadian, Ahmad Al-Dahle, Aiesha Letman, Akhil Mathur, Alan Schelten, Alex Vaughan, and 1 others. 2024.
\newblock The llama 3 herd of models.
\newblock \emph{arXiv preprint arXiv:2407.21783}.

\bibitem[{He et~al.(2025)He, Dai, He, Liu, Tang, Lu, Li, Ding, Mukherjee, Wang et~al.}]{he2025traject}
Pengfei He, Zhenwei Dai, Bing He, Hui Liu, Xianfeng Tang, Hanqing Lu, Juanhui Li, Jiayuan Ding, Subhabrata Mukherjee, Suhang Wang, and 1 others. 2025.
\newblock Traject-bench: A trajectory-aware benchmark for evaluating agentic tool use.
\newblock \emph{arXiv preprint arXiv:2510.04550}.

\bibitem[{Huang et~al.(2024)Huang, Shi, Li, Fan, Wu, Zhang, Liu, Zhou, Wan, Gong et~al.}]{huang2024metatool}
Yue Huang, Jiawen Shi, Yuan Li, Chenrui Fan, Siyuan Wu, Qihui Zhang, Yixin Liu, Pan Zhou, Yao Wan, Neil~Zhenqiang Gong, and 1 others. 2024.
\newblock Metatool benchmark for large language models: Deciding whether to use tools and which to use.
\newblock In \emph{ICLR}.

\bibitem[{Jiang and Ferraro(2024)}]{jiang2024memorization}
Yuxuan Jiang and Francis Ferraro. 2024.
\newblock Memorization over reasoning? exposing and mitigating verbatim memorization in large language models' character understanding evaluation.
\newblock \emph{EACL 2026 Main (Oral)}.

\bibitem[{Jiang and Ferraro(2026)}]{jiang2026scribe}
Yuxuan Jiang and Francis Ferraro. 2026.
\newblock Scribe: Structured mid-level supervision for tool-using language models.
\newblock \emph{arXiv preprint arXiv:2601.03555}.

\bibitem[{Jiang et~al.(2025)Jiang, Li, and Ferraro}]{jiang2025drp}
Yuxuan Jiang, Dawei Li, and Frank Ferraro. 2025.
\newblock Drp: Distilled reasoning pruning with skill-aware step decomposition for efficient large reasoning models.
\newblock \emph{arXiv preprint arXiv:2505.13975}.

\bibitem[{Lee et~al.(2025)Lee, Yang, Baik, Liu, Tan, Li, Wen, Hou, Duong-Tran, Chen et~al.}]{lee2025knowledge}
Joseph Lee, Shu Yang, Jae~Young Baik, Xiaoxi Liu, Zhen Tan, Dawei Li, Zixuan Wen, Bojian Hou, Duy Duong-Tran, Tianlong Chen, and 1 others. 2025.
\newblock Knowledge-driven feature selection and engineering for genotype data with large language models.
\newblock \emph{AMIA Summits on Translational Science Proceedings}, 2025:250.

\bibitem[{Li et~al.(2025{\natexlab{a}})Li, Jiang, Huang, Beigi, Zhao, Tan, Bhattacharjee, Jiang, Chen, Wu et~al.}]{li2025generation}
Dawei Li, Bohan Jiang, Liangjie Huang, Alimohammad Beigi, Chengshuai Zhao, Zhen Tan, Amrita Bhattacharjee, Yuxuan Jiang, Canyu Chen, Tianhao Wu, and 1 others. 2025{\natexlab{a}}.
\newblock From generation to judgment: Opportunities and challenges of llm-as-a-judge.
\newblock In \emph{Proceedings of the 2025 Conference on Empirical Methods in Natural Language Processing}, pages 2757--2791.

\bibitem[{Li et~al.(2025{\natexlab{b}})Li, Zhang, Zhang, Zhang, Liu, Yao, Xu, Zheng, Wang, Chen et~al.}]{li2025system}
Zhong-Zhi Li, Duzhen Zhang, Ming-Liang Zhang, Jiaxin Zhang, Zengyan Liu, Yuxuan Yao, Haotian Xu, Junhao Zheng, Pei-Jie Wang, Xiuyi Chen, and 1 others. 2025{\natexlab{b}}.
\newblock From system 1 to system 2: A survey of reasoning large language models.
\newblock \emph{arXiv preprint arXiv:2502.17419}.

\bibitem[{Liang et~al.(2025)Liang, Li, Gong, Wang, Zhang, Shen, Wu, and Chen}]{liang2025sws}
Xiao Liang, Zhong-Zhi Li, Yeyun Gong, Yang Wang, Hengyuan Zhang, Yelong Shen, Ying~Nian Wu, and Weizhu Chen. 2025.
\newblock Sws: Self-aware weakness-driven problem synthesis in reinforcement learning for llm reasoning.
\newblock \emph{arXiv preprint arXiv:2506.08989}.

\bibitem[{Lightman et~al.(2023)Lightman, Kosaraju, Burda, Edwards, Baker, Lee, Leike, Schulman, Sutskever, and Cobbe}]{lightman2023let}
Hunter Lightman, Vineet Kosaraju, Yuri Burda, Harrison Edwards, Bowen Baker, Teddy Lee, Jan Leike, John Schulman, Ilya Sutskever, and Karl Cobbe. 2023.
\newblock Let's verify step by step.
\newblock In \emph{The Twelfth International Conference on Learning Representations}.

\bibitem[{Lu et~al.(2025)Lu, Holleis, Zhang, Aumayer, Nan, Bai, Ma, Ma, Li, Yin et~al.}]{lu2025toolsandbox}
Jiarui Lu, Thomas Holleis, Yizhe Zhang, Bernhard Aumayer, Feng Nan, Haoping Bai, Shuang Ma, Shen Ma, Mengyu Li, Guoli Yin, and 1 others. 2025.
\newblock Toolsandbox: A stateful, conversational, interactive evaluation benchmark for llm tool use capabilities.
\newblock In \emph{Findings of the Association for Computational Linguistics: NAACL 2025}, pages 1160--1183.

\bibitem[{L{\`u} et~al.(2025)L{\`u}, Kazemnejad, Meade, Patel, Shin, Zambrano, Sta{\'n}czak, Shaw, Pal, and Reddy}]{lu2025agentrewardbench}
Xing~Han L{\`u}, Amirhossein Kazemnejad, Nicholas Meade, Arkil Patel, Dongchan Shin, Alejandra Zambrano, Karolina Sta{\'n}czak, Peter Shaw, Christopher~J Pal, and Siva Reddy. 2025.
\newblock Agentrewardbench: Evaluating automatic evaluations of web agent trajectories.
\newblock \emph{arXiv preprint arXiv:2504.08942}.

\bibitem[{Men et~al.(2025)Men, Jin, Cao, Chen, Liu, and Zhao}]{men-etal-2025-agent}
Tianyi Men, Zhuoran Jin, Pengfei Cao, Yubo Chen, Kang Liu, and Jun Zhao. 2025.
\newblock \href {https://doi.org/10.18653/v1/2025.acl-long.857} {Agent-{R}eward{B}ench: Towards a unified benchmark for reward modeling across perception, planning, and safety in real-world multimodal agents}.
\newblock In \emph{Proceedings of the 63rd Annual Meeting of the Association for Computational Linguistics (Volume 1: Long Papers)}, pages 17521--17541, Vienna, Austria. Association for Computational Linguistics.

\bibitem[{Ning et~al.(2024)Ning, Su, Lv, Zhang, Liu, Liu, and Xu}]{ning2024wtu}
Kangyun Ning, Yisong Su, Xueqiang Lv, Yuanzhe Zhang, Jian Liu, Kang Liu, and Jinan Xu. 2024.
\newblock Wtu-eval: A whether-or-not tool usage evaluation benchmark for large language models.
\newblock \emph{arXiv preprint arXiv:2407.12823}.

\bibitem[{{OpenAI}(2025)}]{openai2025gpt5}
{OpenAI}. 2025.
\newblock \href {https://openai.com/index/introducing-gpt-5/} {Introducing gpt-5}.
\newblock \url{https://openai.com/index/introducing-gpt-5/}.
\newblock Official announcement of OpenAI’s GPT-5, a unified and state-of-the-art AI system available August 7, 2025.

\bibitem[{OpenAI et~al.(2024)OpenAI, Achiam, Adler, Agarwal, Ahmad, Akkaya, Aleman, Almeida, Altenschmidt, Altman, Anadkat, Avila, Babuschkin, Balaji, Balcom, Baltescu, Bao, Bavarian, Belgum, Bello, Berdine, Bernadett-Shapiro, Berner, Bogdonoff, Boiko, Boyd, Brakman, Brockman, Brooks, Brundage, Button, Cai, Campbell, Cann, Carey, Carlson, Carmichael, Chan, Chang, Chantzis, Chen, Chen, Chen, Chen, Chen, Chess, Cho, Chu, Chung, Cummings, Currier, Dai, Decareaux, Degry, Deutsch, Deville, Dhar, Dohan, Dowling, Dunning, Ecoffet, Eleti, Eloundou, Farhi, Fedus, Felix, Fishman, Forte, Fulford, Gao, Georges, Gibson, Goel, Gogineni, Goh, Gontijo-Lopes, Gordon, Grafstein, Gray, Greene, Gross, Gu, Guo, Hallacy, Han, Harris, He, Heaton, Heidecke, Hesse, Hickey, Hickey, Hoeschele, Houghton, Hsu, Hu, Hu, Huizinga, Jain, Jain, Jang, Jiang, Jiang, Jin, Jin, Jomoto, Jonn, Jun, Kaftan, Łukasz Kaiser, Kamali, Kanitscheider, Keskar, Khan, Kilpatrick, Kim, Kim, Kim, Kirchner, Kiros, Knight, Kokotajlo, Łukasz Kondraciuk,
  Kondrich, Konstantinidis, Kosic, Krueger, Kuo, Lampe, Lan, Lee, Leike, Leung, Levy, Li, Lim, Lin, Lin, Litwin, Lopez, Lowe, Lue, Makanju, Malfacini, Manning, Markov, Markovski, Martin, Mayer, Mayne, McGrew, McKinney, McLeavey, McMillan, McNeil, Medina, Mehta, Menick, Metz, Mishchenko, Mishkin, Monaco, Morikawa, Mossing, Mu, Murati, Murk, Mély, Nair, Nakano, Nayak, Neelakantan, Ngo, Noh, Ouyang, O'Keefe, Pachocki, Paino, Palermo, Pantuliano, Parascandolo, Parish, Parparita, Passos, Pavlov, Peng, Perelman, de~Avila Belbute~Peres, Petrov, de~Oliveira~Pinto, Michael, Pokorny, Pokrass, Pong, Powell, Power, Power, Proehl, Puri, Radford, Rae, Ramesh, Raymond, Real, Rimbach, Ross, Rotsted, Roussez, Ryder, Saltarelli, Sanders, Santurkar, Sastry, Schmidt, Schnurr, Schulman, Selsam, Sheppard, Sherbakov, Shieh, Shoker, Shyam, Sidor, Sigler, Simens, Sitkin, Slama, Sohl, Sokolowsky, Song, Staudacher, Such, Summers, Sutskever, Tang, Tezak, Thompson, Tillet, Tootoonchian, Tseng, Tuggle, Turley, Tworek, Uribe, Vallone,
  Vijayvergiya, Voss, Wainwright, Wang, Wang, Wang, Ward, Wei, Weinmann, Welihinda, Welinder, Weng, Weng, Wiethoff, Willner, Winter, Wolrich, Wong, Workman, Wu, Wu, Wu, Xiao, Xu, Yoo, Yu, Yuan, Zaremba, Zellers, Zhang, Zhang, Zhao, Zheng, Zhuang, Zhuk, and Zoph}]{openai2024gpt4technicalreport}
OpenAI, Josh Achiam, Steven Adler, Sandhini Agarwal, Lama Ahmad, Ilge Akkaya, Florencia~Leoni Aleman, Diogo Almeida, Janko Altenschmidt, Sam Altman, Shyamal Anadkat, Red Avila, Igor Babuschkin, Suchir Balaji, Valerie Balcom, Paul Baltescu, Haiming Bao, Mohammad Bavarian, Jeff Belgum, and 262 others. 2024.
\newblock \href {https://arxiv.org/abs/2303.08774} {Gpt-4 technical report}.
\newblock \emph{Preprint}, arXiv:2303.08774.

\bibitem[{Patil et~al.()Patil, Mao, Yan, Ji, Suresh, Stoica, and Gonzalez}]{patilberkeley}
Shishir~G Patil, Huanzhi Mao, Fanjia Yan, Charlie Cheng-Jie Ji, Vishnu Suresh, Ion Stoica, and Joseph~E Gonzalez.
\newblock The berkeley function calling leaderboard (bfcl): From tool use to agentic evaluation of large language models.
\newblock In \emph{Forty-second International Conference on Machine Learning}.

\bibitem[{Qian et~al.(2025)Qian, Acikgoz, He, Wang, Chen, Hakkani-T{\"u}r, Tur, and Ji}]{qian2025toolrl}
Cheng Qian, Emre~Can Acikgoz, Qi~He, Hongru Wang, Xiusi Chen, Dilek Hakkani-T{\"u}r, Gokhan Tur, and Heng Ji. 2025.
\newblock Toolrl: Reward is all tool learning needs.
\newblock \emph{arXiv preprint arXiv:2504.13958}.

\bibitem[{Qin et~al.(2025)Qin, Chen, Zhou, Chen, Li, Liao, Li, Che, and Yu}]{qin2025survey}
Libo Qin, Qiguang Chen, Yuhang Zhou, Zhi Chen, Yinghui Li, Lizi Liao, Min Li, Wanxiang Che, and Philip~S Yu. 2025.
\newblock A survey of multilingual large language models.
\newblock \emph{Patterns}, 6(1).

\bibitem[{Qwen et~al.(2025)Qwen, :, Yang, Yang, Zhang, Hui, Zheng, Yu, Li, Liu, Huang, Wei, Lin, Yang, Tu, Zhang, Yang, Yang, Zhou, Lin, Dang, Lu, Bao, Yang, Yu, Li, Xue, Zhang, Zhu, Men, Lin, Li, Tang, Xia, Ren, Ren, Fan, Su, Zhang, Wan, Liu, Cui, Zhang, and Qiu}]{qwen2025qwen25technicalreport}
Qwen, :, An~Yang, Baosong Yang, Beichen Zhang, Binyuan Hui, Bo~Zheng, Bowen Yu, Chengyuan Li, Dayiheng Liu, Fei Huang, Haoran Wei, Huan Lin, Jian Yang, Jianhong Tu, Jianwei Zhang, Jianxin Yang, Jiaxi Yang, Jingren Zhou, and 25 others. 2025.
\newblock \href {https://arxiv.org/abs/2412.15115} {Qwen2.5 technical report}.
\newblock \emph{Preprint}, arXiv:2412.15115.

\bibitem[{Ren et~al.(2025)Ren, Shao, Song, Xin, Wang, Zhao, Zhang, Fu, Zhu, Yang et~al.}]{ren2025deepseek}
ZZ~Ren, Zhihong Shao, Junxiao Song, Huajian Xin, Haocheng Wang, Wanjia Zhao, Liyue Zhang, Zhe Fu, Qihao Zhu, Dejian Yang, and 1 others. 2025.
\newblock Deepseek-prover-v2: Advancing formal mathematical reasoning via reinforcement learning for subgoal decomposition.
\newblock \emph{arXiv preprint arXiv:2504.21801}.

\bibitem[{Shahroz et~al.(2025)Shahroz, Tan, Yun, Fleming, and Chen}]{shahroz-etal-2025-agents}
Rana Shahroz, Zhen Tan, Sukwon Yun, Charles Fleming, and Tianlong Chen. 2025.
\newblock \href {https://doi.org/10.18653/v1/2025.acl-long.476} {Agents under siege: Breaking pragmatic multi-agent {LLM} systems with optimized prompt attacks}.
\newblock In \emph{Proceedings of the 63rd Annual Meeting of the Association for Computational Linguistics (Volume 1: Long Papers)}, pages 9661--9674, Vienna, Austria. Association for Computational Linguistics.

\bibitem[{Shao et~al.(2024)Shao, Wang, Zhu, Xu, Song, Bi, Zhang, Zhang, Li, Wu et~al.}]{shao2024deepseekmath}
Zhihong Shao, Peiyi Wang, Qihao Zhu, Runxin Xu, Junxiao Song, Xiao Bi, Haowei Zhang, Mingchuan Zhang, YK~Li, Yang Wu, and 1 others. 2024.
\newblock Deepseekmath: Pushing the limits of mathematical reasoning in open language models.
\newblock \emph{arXiv preprint arXiv:2402.03300}.

\bibitem[{Shen(2024)}]{shen2024llm}
Zhuocheng Shen. 2024.
\newblock Llm with tools: A survey.
\newblock \emph{arXiv preprint arXiv:2409.18807}.

\bibitem[{Snell et~al.(2024)Snell, Lee, Xu, and Kumar}]{snell2024scaling}
Charlie Snell, Jaehoon Lee, Kelvin Xu, and Aviral Kumar. 2024.
\newblock Scaling llm test-time compute optimally can be more effective than scaling model parameters.
\newblock \emph{arXiv preprint arXiv:2408.03314}.

\bibitem[{Song et~al.(2025)Song, Su, Qu, Zhou, and Cheng}]{song-etal-2025-prmbench}
Mingyang Song, Zhaochen Su, Xiaoye Qu, Jiawei Zhou, and Yu~Cheng. 2025.
\newblock \href {https://doi.org/10.18653/v1/2025.acl-long.1230} {{PRMB}ench: A fine-grained and challenging benchmark for process-level reward models}.
\newblock In \emph{Proceedings of the 63rd Annual Meeting of the Association for Computational Linguistics (Volume 1: Long Papers)}, pages 25299--25346, Vienna, Austria. Association for Computational Linguistics.

\bibitem[{Stiennon et~al.(2020)Stiennon, Ouyang, Wu, Ziegler, Lowe, Voss, Radford, Amodei, and Christiano}]{stiennon2020learning}
Nisan Stiennon, Long Ouyang, Jeffrey Wu, Daniel Ziegler, Ryan Lowe, Chelsea Voss, Alec Radford, Dario Amodei, and Paul~F Christiano. 2020.
\newblock Learning to summarize with human feedback.
\newblock \emph{Advances in neural information processing systems}, 33:3008--3021.

\bibitem[{Sun et~al.(2024)Sun, Yu, Shen, Liu, Yang, Welleck, and Gan}]{sun2024easy}
Zhiqing Sun, Longhui Yu, Yikang Shen, Weiyang Liu, Yiming Yang, Sean Welleck, and Chuang Gan. 2024.
\newblock Easy-to-hard generalization: Scalable alignment beyond human supervision.
\newblock \emph{arXiv preprint arXiv:2403.09472}.

\bibitem[{Tan et~al.(2024{\natexlab{a}})Tan, Cheng, Wang, Yuan, Li, and Liu}]{tan2024interpreting}
Zhen Tan, Lu~Cheng, Song Wang, Bo~Yuan, Jundong Li, and Huan Liu. 2024{\natexlab{a}}.
\newblock Interpreting pretrained language models via concept bottlenecks.
\newblock In \emph{Pacific-Asia Conference on Knowledge Discovery and Data Mining}, pages 56--74. Springer.

\bibitem[{Tan et~al.(2024{\natexlab{b}})Tan, Li, Wang, Beigi, Jiang, Bhattacharjee, Karami, Li, Cheng, and Liu}]{tan2024large}
Zhen Tan, Dawei Li, Song Wang, Alimohammad Beigi, Bohan Jiang, Amrita Bhattacharjee, Mansooreh Karami, Jundong Li, Lu~Cheng, and Huan Liu. 2024{\natexlab{b}}.
\newblock Large language models for data annotation and synthesis: A survey.
\newblock \emph{arXiv preprint arXiv:2402.13446}.

\bibitem[{Tan et~al.(2025)Tan, Yan, Hsu, Han, Wang, Le, Song, Chen, Palangi, Lee et~al.}]{tan2025prospect}
Zhen Tan, Jun Yan, I-Hung Hsu, Rujun Han, Zifeng Wang, Long Le, Yiwen Song, Yanfei Chen, Hamid Palangi, George Lee, and 1 others. 2025.
\newblock In prospect and retrospect: Reflective memory management for long-term personalized dialogue agents.
\newblock In \emph{Proceedings of the 63rd Annual Meeting of the Association for Computational Linguistics (Volume 1: Long Papers)}, pages 8416--8439.

\bibitem[{Uesato et~al.(2022)Uesato, Kushman, Kumar, Song, Siegel, Wang, Creswell, Irving, and Higgins}]{uesato2022solving}
Jonathan Uesato, Nate Kushman, Ramana Kumar, Francis Song, Noah Siegel, Lisa Wang, Antonia Creswell, Geoffrey Irving, and Irina Higgins. 2022.
\newblock Solving math word problems with process-and outcome-based feedback.
\newblock \emph{arXiv preprint arXiv:2211.14275}.

\bibitem[{von Werra et~al.(2020)von Werra, Belkada, Tunstall, Beeching, Thrush, Lambert, Huang, Rasul, and Gallouédec}]{vonwerra2022trl}
Leandro von Werra, Younes Belkada, Lewis Tunstall, Edward Beeching, Tristan Thrush, Nathan Lambert, Shengyi Huang, Kashif Rasul, and Quentin Gallouédec. 2020.
\newblock Trl: Transformer reinforcement learning.
\newblock \url{https://github.com/huggingface/trl}.

\bibitem[{Wang et~al.(2024{\natexlab{a}})Wang, Zerun, Li, Zhang, Chen, Chen, and Le}]{wang2024gta}
Jize Wang, Ma~Zerun, Yining Li, Songyang Zhang, Cailian Chen, Kai Chen, and Xinyi Le. 2024{\natexlab{a}}.
\newblock Gta: a benchmark for general tool agents.
\newblock \emph{Advances in Neural Information Processing Systems}, 37:75749--75790.

\bibitem[{Wang et~al.(2024{\natexlab{b}})Wang, Li, Shao, Xu, Dai, Li, Chen, Wu, and Sui}]{wang2024math}
Peiyi Wang, Lei Li, Zhihong Shao, Runxin Xu, Damai Dai, Yifei Li, Deli Chen, Yu~Wu, and Zhifang Sui. 2024{\natexlab{b}}.
\newblock Math-shepherd: Verify and reinforce llms step-by-step without human annotations.
\newblock In \emph{Proceedings of the 62nd Annual Meeting of the Association for Computational Linguistics (Volume 1: Long Papers)}, pages 9426--9439.

\bibitem[{Wang et~al.(2024{\natexlab{c}})Wang, Tong, Zhang, Li, Zhang, and Chen}]{wang2024bpo}
Sizhe Wang, Yongqi Tong, Hengyuan Zhang, Dawei Li, Xin Zhang, and Tianlong Chen. 2024{\natexlab{c}}.
\newblock Bpo: Towards balanced preference optimization between knowledge breadth and depth in alignment.
\newblock \emph{arXiv preprint arXiv:2411.10914}.

\bibitem[{Wang et~al.(2024{\natexlab{d}})Wang, Kulikov, Golovneva, Yu, Yuan, Dwivedi-Yu, Pang, Fazel-Zarandi, Weston, and Li}]{wang2024self}
Tianlu Wang, Ilia Kulikov, Olga Golovneva, Ping Yu, Weizhe Yuan, Jane Dwivedi-Yu, Richard~Yuanzhe Pang, Maryam Fazel-Zarandi, Jason Weston, and Xian Li. 2024{\natexlab{d}}.
\newblock Self-taught evaluators.
\newblock \emph{arXiv preprint arXiv:2408.02666}.

\bibitem[{Wang et~al.()Wang, Wei, Schuurmans, Le, Chi, Narang, Chowdhery, and Zhou}]{wangself}
Xuezhi Wang, Jason Wei, Dale Schuurmans, Quoc~V Le, Ed~H Chi, Sharan Narang, Aakanksha Chowdhery, and Denny Zhou.
\newblock Self-consistency improves chain of thought reasoning in language models.
\newblock In \emph{The Eleventh International Conference on Learning Representations}.

\bibitem[{Xu et~al.(2025)Xu, Jiang, Dipta, and Hengyuan}]{xu2025learning}
Ningning Xu, Yuxuan Jiang, Shubhashis~Roy Dipta, and Zhang Hengyuan. 2025.
\newblock Learning how to use tools, not just when: Pattern-aware tool-integrated reasoning.
\newblock \emph{MATH-AI @ NeurIPS 2025}.

\bibitem[{Yang et~al.(2025{\natexlab{a}})Yang, Li, Yang, Zhang, Hui, Zheng, Yu, Gao, Huang, Lv et~al.}]{yang2025qwen3}
An~Yang, Anfeng Li, Baosong Yang, Beichen Zhang, Binyuan Hui, Bo~Zheng, Bowen Yu, Chang Gao, Chengen Huang, Chenxu Lv, and 1 others. 2025{\natexlab{a}}.
\newblock Qwen3 technical report.
\newblock \emph{arXiv preprint arXiv:2505.09388}.

\bibitem[{Yang et~al.(2024)Yang, Zhang, Hui, Gao, Yu, Li, Liu, Tu, Zhou, Lin et~al.}]{yang2024qwen2}
An~Yang, Beichen Zhang, Binyuan Hui, Bofei Gao, Bowen Yu, Chengpeng Li, Dayiheng Liu, Jianhong Tu, Jingren Zhou, Junyang Lin, and 1 others. 2024.
\newblock Qwen2. 5-math technical report: Toward mathematical expert model via self-improvement.
\newblock \emph{arXiv preprint arXiv:2409.12122}.

\bibitem[{Yang et~al.(2025{\natexlab{b}})Yang, Wu, Ding, Wu, Liang, Gong, Zhang, and Zhang}]{yang2025quantifying}
Shiping Yang, Jie Wu, Wenbiao Ding, Ning Wu, Shining Liang, Ming Gong, Hengyuan Zhang, and Dongmei Zhang. 2025{\natexlab{b}}.
\newblock Quantifying the robustness of retrieval-augmented language models against spurious features in grounding data.
\newblock \emph{arXiv preprint arXiv:2503.05587}.

\bibitem[{Yao et~al.(2022)Yao, Zhao, Yu, Du, Shafran, Narasimhan, and Cao}]{yao2022react}
Shunyu Yao, Jeffrey Zhao, Dian Yu, Nan Du, Izhak Shafran, Karthik~R Narasimhan, and Yuan Cao. 2022.
\newblock React: Synergizing reasoning and acting in language models.
\newblock In \emph{The eleventh international conference on learning representations}.

\bibitem[{Ye et~al.(2025)Ye, Du, Yao, Lin, Xu, Chen, Wang, Zhu, Xi, Yuan et~al.}]{ye2025toolhop}
Junjie Ye, Zhengyin Du, Xuesong Yao, Weijian Lin, Yufei Xu, Zehui Chen, Zaiyuan Wang, Sining Zhu, Zhiheng Xi, Siyu Yuan, and 1 others. 2025.
\newblock Toolhop: A query-driven benchmark for evaluating large language models in multi-hop tool use.
\newblock In \emph{Proceedings of the 63rd Annual Meeting of the Association for Computational Linguistics (Volume 1: Long Papers)}, pages 2995--3021.

\bibitem[{Yu et~al.(2025)Yu, Zhang, Zhang, Liang, Zhang, Zhang, Yang, Khademi, Awadalla, Wang et~al.}]{yu2025chain}
Yiyao Yu, Yuxiang Zhang, Dongdong Zhang, Xiao Liang, Hengyuan Zhang, Xingxing Zhang, Ziyi Yang, Mahmoud Khademi, Hany Awadalla, Junjie Wang, and 1 others. 2025.
\newblock Chain-of-reasoning: Towards unified mathematical reasoning in large language models via a multi-paradigm perspective.
\newblock \emph{arXiv preprint arXiv:2501.11110}.

\bibitem[{Zhang et~al.(2025{\natexlab{a}})Zhang, Chen, Qiu, Liang, Li, Wang, Li, Mo, So, and Wong}]{zhang2025guilomoallocatingexpertnumber}
Hengyuan Zhang, Xinrong Chen, Yingmin Qiu, Xiao Liang, Ziyue Li, Guanyu Wang, Weiping Li, Tong Mo, Hayden Kwok-Hay So, and Ngai Wong. 2025{\natexlab{a}}.
\newblock \href {https://arxiv.org/abs/2506.14646} {Guilomo: Allocating expert number and rank for lora-moe via bilevel optimization with guidedselection vectors}.
\newblock \emph{Preprint}, arXiv:2506.14646.

\bibitem[{Zhang et~al.(2025{\natexlab{b}})Zhang, Shang, Wang, Zhang, Yu, Yao, Sun, Yang, and Wei}]{zhang-etal-2025-shifcon}
Hengyuan Zhang, Chenming Shang, Sizhe Wang, Dongdong Zhang, Yiyao Yu, Feng Yao, Renliang Sun, Yujiu Yang, and Furu Wei. 2025{\natexlab{b}}.
\newblock \href {https://doi.org/10.18653/v1/2025.acl-long.239} {{S}hif{C}on: Enhancing non-dominant language capabilities with a shift-based multilingual contrastive framework}.
\newblock In \emph{Proceedings of the 63rd Annual Meeting of the Association for Computational Linguistics (Volume 1: Long Papers)}, pages 4818--4841, Vienna, Austria. Association for Computational Linguistics.

\bibitem[{Zhang et~al.(2025{\natexlab{c}})Zhang, Yang, Liang, Shang, Jiang, Tao, Xiong, So, Xie, Chang et~al.}]{zhang2025find}
Hengyuan Zhang, Shiping Yang, Xiao Liang, Chenming Shang, Yuxuan Jiang, Chaofan Tao, Jing Xiong, Hayden Kwok-Hay So, Ruobing Xie, Angel~X Chang, and 1 others. 2025{\natexlab{c}}.
\newblock Find your optimal teacher: Personalized data synthesis via router-guided multi-teacher distillation.
\newblock \emph{arXiv preprint arXiv:2510.10925}.

\bibitem[{Zhang et~al.(2025{\natexlab{d}})Zhang, Qiu, Tan, Zhang, Lu, Peng, Xu, Agudelo, Qian, and Chen}]{zhang2025symbiotic}
Ruichen Zhang, Mufan Qiu, Zhen Tan, Mohan Zhang, Vincent Lu, Jie Peng, Kaidi Xu, Leandro~Z Agudelo, Peter Qian, and Tianlong Chen. 2025{\natexlab{d}}.
\newblock Symbiotic cooperation for web agents: Harnessing complementary strengths of large and small llms.
\newblock \emph{arXiv preprint arXiv:2502.07942}.

\bibitem[{Zhao et~al.(2025{\natexlab{a}})Zhao, Tan, Wong, Zhao, Chen, and Liu}]{zhao2025scale}
Chengshuai Zhao, Zhen Tan, Chau-Wai Wong, Xinyan Zhao, Tianlong Chen, and Huan Liu. 2025{\natexlab{a}}.
\newblock Scale: Towards collaborative content analysis in social science with large language model agents and human intervention.
\newblock \emph{arXiv preprint arXiv:2502.10937}.

\bibitem[{Zhao et~al.(2025{\natexlab{b}})Zhao, Liu, Zhang, Zhou, Gao, Li, Lyu, Qian, Qi, Li et~al.}]{zhao2025genprm}
Jian Zhao, Runze Liu, Kaiyan Zhang, Zhimu Zhou, Junqi Gao, Dong Li, Jiafei Lyu, Zhouyi Qian, Biqing Qi, Xiu Li, and 1 others. 2025{\natexlab{b}}.
\newblock Genprm: Scaling test-time compute of process reward models via generative reasoning.
\newblock \emph{arXiv preprint arXiv:2504.00891}.

\bibitem[{Zheng et~al.(2024)Zheng, Zhang, Zhang, Ye, Luo, Feng, and Ma}]{zheng2024llamafactory}
Yaowei Zheng, Richong Zhang, Junhao Zhang, Yanhan Ye, Zheyan Luo, Zhangchi Feng, and Yongqiang Ma. 2024.
\newblock \href {http://arxiv.org/abs/2403.13372} {Llamafactory: Unified efficient fine-tuning of 100+ language models}.
\newblock In \emph{Proceedings of the 62nd Annual Meeting of the Association for Computational Linguistics (Volume 3: System Demonstrations)}, Bangkok, Thailand. Association for Computational Linguistics.

\bibitem[{Zhou et~al.(2024)Zhou, Yan, Shlapentokh-Rothman, Wang, and Wang}]{zhou2024language}
Andy Zhou, Kai Yan, Michal Shlapentokh-Rothman, Haohan Wang, and Yu-Xiong Wang. 2024.
\newblock Language agent tree search unifies reasoning, acting, and planning in language models.
\newblock In \emph{Proceedings of the 41st International Conference on Machine Learning}, pages 62138--62160.

\end{thebibliography}

\clearpage
\appendix

\section{More Details of ToolPRMBench}
\label{app:ToolPRMBench}

Table~\ref{tab:dataset_stats} presents the details of the training-testing split in ToolPRMBench. We also show the prompt template we use in LLM-based annotation and multi-LLM verification in the tables below.

\begin{table}[htbp]
\centering
\small
\begin{tabular}{lccc}
\toprule[1.5pt]
\textbf{Dataset} & \textbf{\#train} & \textbf{\#test} & \textbf{\#all} \\ \midrule
BFCL             & 243              & 111             & 354            \\
ToolSandbox      & 299              & 130             & 429            \\
GTA              & 0                & 118             & 118            \\
ToolTalk         & 0                & 86              & 86             \\ \midrule
\textbf{all}     & \textbf{542}     & \textbf{445}    & \textbf{987}   \\ 
\bottomrule[1.5pt]
\end{tabular}
\caption{Details statistics of ToolPRMBench.}
\label{tab:dataset_stats}
\end{table}

\begin{tcolorbox}[breakable, title={PRM Annotation Prompt for ToolSandbox}, label=prompt:prm_general]
\small
\ttfamily
You are an annotator LLM for building a process reward model benchmark.\\
\\
Given an INPUT\_JSON with: sample\_id, trajectory, available\_tools, milestones, milestone\_edges, and raw tool feedback.\\
\\
Do: Identify the first incorrect agent step. A step is incorrect if it violates ANY of:
\begin{itemize}
    \item wrong tool choice,
    \item wrong tool parameters,
    \item incorrect user-facing response,
    \item skipped or unordered milestone,
    \item performing a later milestone before an earlier required one.
\end{itemize}
Output a single JSON object (no extra text) with this exact schema:\\
\{\\
  "sample\_id": "<string>",\\
  "history": [...],\\
  "action\_rejected": <the incorrect agent action>,\\
  "action\_chosen": <the corrected action>,\\
  "rationale": "<short, concrete explanation>",\\
  "error type": "<short phrase>"\\
\}
\end{tcolorbox}
\begin{tcolorbox}[breakable, title={PRM Annotation Prompt for BFCL}, label=prompt:prm_bfcl]
\small
\ttfamily
You are an annotator LLM for building process reward model benchmark samples.\\
\\
Each input is a JSON object with keys: id, model\_name, test\_category, valid, error, execution\_result, prompt, possible\_answer, history, and function.\\
\\
Your Task: Return one JSON object (no extra text) with this schema:\\
\{\\
  "sample\_id": "<same as input>",\\
  "history": [...],\\
  "action\_rejected": \{ "name": "<...>", "parameters": \{ ... \} \},\\
  "action\_chosen": \{ "name": "<...>", "parameters": \{ ... \} \},\\
  "rationale": "1--2 sentences explaining the judgment.",\\
  "error type": "<short phrase>"\\
\}
\end{tcolorbox}

\section{More Details of Experiment Implementation}
\label{app:Implementation}

\subsection{Training Environment and Infrastructure}
All experiments are conducted on a single machine equipped with 8 $\times$ NVIDIA H20 (96GB) GPUs. We utilize the DeepSpeed library to optimize training efficiency. Specifically, \textbf{DeepSpeed ZeRO-3} is employed to shard model states, gradients, and optimizer states across all 8 GPUs, enabling full-parameter fine-tuning of the backbone model within the available memory budget.

\subsection{Supervised Fine-Tuning (SFT)}
We use \textbf{Qwen3-4B} as our base model. The SFT stage is implemented using the LLaMA-Factory framework. We perform full-parameter fine-tuning for 2 epochs using the AdamW optimizer. The learning rate is set to $1.0 \times 10^{-5}$ with a \textbf{cosine} scheduler and a warmup ratio of 0.1. To accommodate long-context reasoning in Chain-of-Thought (CoT) tasks, the \textbf{cutoff length} is set to 4,096 tokens. For the standard base tasks, we use a per-device batch size of 4. For CoT-enhanced datasets, we reduce the per-device batch size to 1 to manage memory consumption. All SFT experiments are performed in \textbf{bf16} precision.

\subsection{Reinforcement Learning (RL)}
Following SFT, we perform reinforcement learning using the \textbf{Group Relative Policy Optimization (GRPO)} algorithm, implemented via the TRL library. The model is trained for 1 epoch with a peak learning rate of $2.0 \times 10^{-6}$. In each training step, GRPO generates a group of $G=8$ completions per prompt to compute relative rewards. 

The maximum prompt length is 2,048 tokens, and the maximum completion length is 4,096 tokens to provide sufficient space for CoT reasoning. We use a \textbf{KL divergence coefficient} of $\beta=0.01$ to regularize the policy. The reward function is defined based on binary accuracy: a reward of 1.0 is assigned if the ground-truth answer is present in the model's output following the \texttt{</think>} token; otherwise, the reward is 0.0.

\begin{table}[htbp]
\centering
\small
\begin{tabular}{lcc}
\toprule[1.5pt]
\textbf{Parameter} & \textbf{SFT} & \textbf{RL (GRPO)} \\ \midrule
Backbone Model      & Qwen3-4B     & Qwen3-4B (SFT)     \\
Hardware            & \multicolumn{2}{c}{1 $\times$ 8-H20 GPU Server} \\
Optimizer           & AdamW        & AdamW              \\
Learning Rate       & $1.0 \times 10^{-5}$ & $2.0 \times 10^{-6}$ \\
LR Scheduler        & Cosine       & Cosine             \\
Training Epochs     & 2.0          & 1.0                \\
Precision           & bf16         & bf16               \\
Max Length          & 4,096        & 4,096              \\
Batch Size / Dev    & 1 / 4        & 4                  \\
Group Size ($G$)    & N/A          & 8                  \\
KL Coeff ($\beta$)  & N/A          & 0.01               \\
DeepSpeed Stage     & ZeRO-3       & ZeRO-3             \\ \bottomrule[1.5pt]
\end{tabular}
\caption{Hyperparameters for SFT and RL (GRPO) training stages.}
\label{tab:hyperparams}
\end{table}

We also present the prompt we used to evaluate various models below, as well as the prompt template we used to distill CoT from the teacher model and to perform data synthesis.

\begin{tcolorbox}[breakable, title={Standard Evaluation Prompt}, label=prompt:run_std]
\small
\ttfamily
Given the function descriptin, history, and the following two actions, which action is the correct one that could help to finish the task:\\
\\
\#\# Function Description: \{functions\}\\
\\
\#\# History: \{history\}\\
\\
\#\# Action1: \{action\_1\}\\
\\
\#\# Action2: \{action\_2\}\\
\\
Please generate your annswer in a JSON format:\\
\{\\
    "rationale": <reasoning process for the judgment>,\\
    "chosen action": <"Action1" or "Action2">\\
\}
\end{tcolorbox}
\begin{tcolorbox}[breakable, title={Evaluation Prompt for ToolPRM-Base and ToolPRM-GRPO}, label=prompt:run_base]
\small
\ttfamily
Given the interaction history, function description and two actions, which action is the correct intermidiate step that could help in finishing the task:\\
\\
\#\# history: \{history\}\\
\\
\#\# function description: \{func\_desc\}\\
\\
\#\# action\_1: \{action\_1\}\\
\\
\#\# action\_2: \{action\_2\}\\
\\
Please only generate action\_1 or action\_2 as the final answer.
\end{tcolorbox}
\begin{tcolorbox}[breakable, title={Evaluation Prompt for ToolPRM-CoT}, label=prompt:run_cot]
\small
\ttfamily
You are an expert evaluator. You are given: interaction history, available tool descriptions, and two candidate assistant actions.\\
\\
Your task is to: Produce a concise chain-of-thought style rationale (2-6 sentences) that explains which of the two actions is better. Also indicate the winning action.\\
\\
Schema:\\
\{\\
  "rationale": "concise stepwise rationale (2-6 sentences)",\\
  "winning\_action": "action\_1|action\_2"\\
\}\\
\\
Now INPUT:\\
available\_tools: \{tool\_desc\}\\
history: \{history\}\\
action\_1: \{action\_1\}\\
action\_2: \{action\_2\}
\end{tcolorbox}
\begin{tcolorbox}[breakable, title={CoT Distillation Prompt}, label=prompt:cot_distillation]
\small
\ttfamily
System: You are an expert evaluator. Produce a concise chain-of-thought rationale and a final judgment.\\
\\
User: You are given:\\
- an interaction history,\\
- available tool / function descriptions,\\
- and two candidate assistant actions (one correct, one incorrect).\\
\\
Your task is to:\\
Produce a concise chain-of-thought style rationale (2-6 sentences) that explains, step by step, why one action is better than the other given the context.\\
Also indicate the winning action as "action\_1" or "action\_2".\\
\\
Schema (you MUST output JSON):\\
\{\\
  "rationale": "concise stepwise rationale (2-6 sentences)",\\
  "winning\_action": "action\_1|action\_2"\\
\}\\
\\
Now INPUT:\\
available\_tools:\\
\{tool\_desc\}\\
\\
history:\\
\{history\_str\}\\
\\
action\_1:\\
\{action\_1\}\\
\\
action\_2:\\
\{action\_2\}
\end{tcolorbox}
\begin{tcolorbox}[breakable, title={Error Injection Prompt}, label=prompt:inject_error]
\small
\ttfamily
Task:\\
You are given the golden conversation history (an array of objects with keys 'role' and 'content') and available tool descriptions (contained in history/system messages).\\
\\
1. Identify a single 'assistant' step where a model is likely to make a mistake (e.g., wrong tool, wrong parameters, hallucinated information, or incorrect response format).\\
2. You MUST produce a mistake of type: \{error\_type\}.\\
3. Construct the 'action\_rejected' (the mistaken version) and 'action\_chosen' (the original correct version).\\
4. Output a valid JSON object only.\\
\\
Schema:\\
\{\\
  "chosen\_index": <int, the 0-based index of the assistant message in history to corrupt>,\\
  "history": <array of messages up to but NOT including the chosen\_index>,\\
  "action\_chosen": <the original content/tool\_call at chosen\_index>,\\
  "action\_rejected": <the new corrupted/mistaken content or tool\_call>,\\
  "rationale": "<short explanation of why this error is plausible>",\\
  "error type": "<short phrase summarizing the error>"\\
\}\\
\\
Golden History:\\
\{history\_str\}
\end{tcolorbox}

\section{The Use of LLMs for Writing}
We employed Google's Gemini 2.5 Pro and OpenAI's GPT-5 as writing assistance tools during the preparation of this manuscript. Their role was exclusively for language refinement, such as improving readability and rephrasing for clarity in an academic writing style. This usage aligns with standard academic practices for language polishing.

\end{document}